\PassOptionsToPackage{table}{xcolor}

\documentclass[sigconf]{acmart}

\copyrightyear{2027}
\acmYear{2027}
\setcopyright{cc}
\setcctype{by}
\acmConference[SenSys '27]{ACM/IEEE International Conference on Embedded Artificial Intelligence and Sensing Systems}{May 17--20, 2027}{Boulder, CO, USA}
\acmBooktitle{ACM/IEEE International Conference on Embedded Artificial Intelligence and Sensing Systems (SenSys '27), May 17--20, 2027, Boulder, CO, USA}
\acmDOI{10.1145/3862575.3862833}
\acmISBN{979-8-4007-3124-2/2027/05}

\AtBeginDocument{%
  }

\usepackage{amsmath,amsfonts,bm}

\def\eqref#1{equation~\ref{#1}}

\def\1{\bm{1}}

\def\mM{{\bm{M}}}

\def\mO{{\bm{O}}}

\def\mR{{\bm{R}}}
\def\mS{{\bm{S}}}

\def\mZ{{\bm{Z}}}

\DeclareMathAlphabet{\mathsfit}{\encodingdefault}{\sfdefault}{m}{sl}
\SetMathAlphabet{\mathsfit}{bold}{\encodingdefault}{\sfdefault}{bx}{n}
\newcommand{\tens}[1]{\bm{\mathcal{#1}}}
\def\tA{{\tens{A}}}

\def\tR{{\tens{R}}}
\def\tS{{\tens{S}}}

\def\tX{{\tens{X}}}

\def\gF{{\mathcal{F}}}

\def\gN{{\mathcal{N}}}

\def\sN{{\mathbb{N}}}

\def\sR{{\mathbb{R}}}

\newcommand{\E}{\mathbb{E}}

\usepackage{subcaption}
\usepackage{mathtools}
\usepackage{amsthm}
\usepackage[capitalize,noabbrev]{cleveref}
\usepackage{algorithm}
\usepackage{algpseudocode}
\usepackage{enumitem}

\theoremstyle{plain}
\newtheorem{theorem}{Theorem}[section]
\newtheorem{proposition}[theorem]{Proposition}

\theoremstyle{definition}

\theoremstyle{remark}

\usepackage[textsize=tiny]{todonotes}

\usepackage{nicefrac}       
\usepackage{wasysym}
\usepackage{wrapfig}
\usepackage{mwe}
\usepackage{multirow}
\usepackage{adjustbox}
\newcommand{\MODEL}{{SPAR}}
\newcommand{\etal}{\textit{et al.}\xspace}
\newcommand{\mask}{\text{\texttt{mask}}}
\newcommand{\concat}{\text{\texttt{cat}}}

\newcommand{\fenc}{\gF_\mathrm{enc}^{(k)}}
\newcommand{\fencnk}{\gF_\mathrm{enc}}

\newcommand{\fjointenc}{\gF_\mathrm{joint\_enc}}
\newcommand{\fsigdec}{\gF_\mathrm{sig\_dec}^{(k)}}
\newcommand{\fsigdecnk}{\gF_\mathrm{sig\_dec}}
\newcommand{\fspdec}{\gF_\mathrm{sp\_dec}^{(k)}}
\newcommand{\fspdecnk}{\gF_\mathrm{sp\_dec}}

\begin{document}

\title[SPAR: Placement-aware Self-supervised Representation Learning for Distributed Sensing]{SPAR: Placement-aware Self-supervised Representation Learning for Distributed Sensing}


\author{Yizhuo Chen}
\author{Tianchen Wang}
\author{You Lyu}
\affiliation{%
  \institution{University of Illinois Urbana-Champaign}
  \city{Urbana}
  \state{IL}
  \country{USA}}
\email{yizhuoc@illinois.edu}

\author{Yanlan Hu}
\affiliation{%
  \institution{Stanford University}
  \city{Stanford}
  \state{CA}
  \country{USA}}
\email{hyl1225@stanford.edu}

\author{Jinyang Li}
\authornote{Corresponding author.}
\author{Tomoyoshi Kimura}
\author{Hongjue Zhao}
\author{Yigong Hu}
\author{Denizhan Kara}
\affiliation{%
  \institution{University of Illinois Urbana-Champaign}
  \city{Urbana}
  \state{IL}
  \country{USA}}
\email{jinyang7@illinois.edu}

\author{Jeffrey Twigg}
\author{Maggie Wigness}
\affiliation{%
  \institution{US Army Research Laboratory}
  \city{Adelphi}
  \state{MD}
  \country{USA}}
\email{jeffrey.n.twigg.civ@army.mil}

\author{Tarek Abdelzaher}
\affiliation{%
  \institution{University of Illinois Urbana-Champaign}
  \city{Urbana}
  \state{IL}
  \country{USA}}
\email{zaher@illinois.edu}

\renewcommand{\shortauthors}{Chen et al.}

\begin{abstract}
To support distributed sensing applications that require understanding of entities in the environment and their locations, this paper develops a self-supervised approach for {\em placement-aware\/} representation learning in spatially-distributed multi-sensor systems. The goal is to produce a latent representation of the physical environment that decouples upstream sensor placement from downstream analytics (in the sense that the pretrained encoder is reused without retraining when sensor placement changes), while supporting spatial reasoning tasks, such as joint target localization and classification, that are sensitive to sensor placement.
A challenge lies in
the {\em multiplicity of sensing vantage points\/} that collectively offer information on the {\em shared
physical environment\/}. The dependence of measurements on sensor locations implies that the latent representation must jointly encode both the observed signals and observer vantage points, doing so in a way that supports the downstream tasks.
Moreover, since the exact deployment-time sensor placement is generally not known at training time, the representation learning framework must be pre-trained to produce such a representation {\em without (training-time) knowledge of deployment-time placement\/}. This challenge distinguishes our goal from approaches, such as NeRF and Gaussian Splatting, that are trained to represent each {\em particular environment given a specific placement of observation points\/} in that environment. In contrast, our representation learning solution, called \MODEL{}, pretrains an encoder independently of future placement by utilizing a novel training objective that emphasizes the duality between signals and vantage points. We offer an interpretation of \MODEL{} from an information-theoretic perspective and show, in the evaluation (based on real-world datasets spanning vehicle monitoring, earthquake localization, and wearable activity recognition), that our representation improves downstream spatial analytics without compromising on other tasks, compared to relying on other representation-learning baselines. A field study further explores \MODEL{} run-time overheads under realistic operating conditions, demonstrating practical feasibility.
\end{abstract}

\ccsdesc[500]{Computing methodologies~Unsupervised learning}
\ccsdesc[300]{Computer systems organization~Sensor networks}

\keywords{self-supervised learning, distributed sensing, representation learning, multi-modal sensor fusion, masked autoencoders}

\maketitle

\section{Introduction}
\label{sec:intro}

\begin{figure*}[!t] \centering
\includegraphics[width=0.8\linewidth]{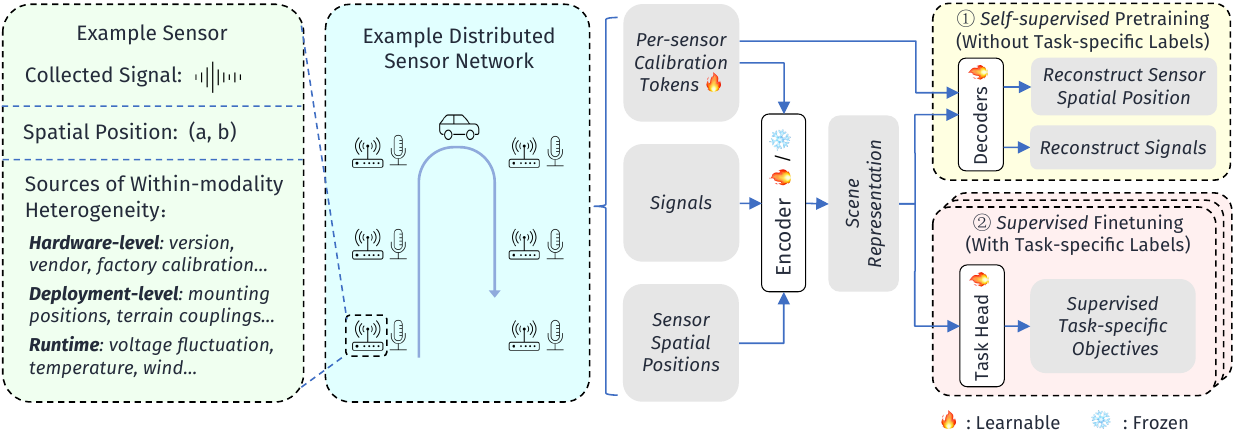} \vspace{-0.2in} \caption{\small Overview of \MODEL{}. Each sensor in a distributed network contributes a signal and a spatial position, and may exhibit within-modality heterogeneity. \MODEL{}'s encoder maps the per-sensor signals, spatial positions, and a learnable per-sensor calibration token into a single representation. \textbf{\textcircled{\small 1} Self-supervised pretraining}, without task-specific labels, trains two decoders to reconstruct sensor signals and spatial positions from that representation. \textbf{\textcircled{\small 2} Supervised fine-tuning}, with task-specific labels, freezes the encoder and trains a lightweight task head on top to perform desired downstream analytics.} \label{fig:motivation}
\vspace{-0.1in}
\end{figure*}

The rise of foundation models (in AI) in the last five years~\cite{bommasani2021opportunities} has placed a growing focus on the importance of self-supervised representation learning~\cite{ericsson2022self} as a means to facilitate downstream data processing. By self-supervised representation learning we refer to the idea that complex data inputs (to an AI model) can be mapped in a self-supervised fashion into an appropriate, well-organized latent space that facilitates downstream perception, reasoning, and decision tasks, hereafter summarily called {\em downstream tasks\/}. Better representation learning can improve downstream task accuracy by retaining (from input data) information that's more relevant to the downstream tasks. The art in {\em self-supervised\/} representation learning lies in ensuring that the produced representation retains the right information {\em without relying on labeled data\/} for training.

The paper advances self-supervised representation learning in IoT applications. A common form of analytics in such applications is to reason about properties and/or locations of entities or events in the environment. Examples include earthquake detection~\citep{caltech1926socal}, vehicle monitoring~\citep{li2025restoreml, bathla2022autonomous}, spatial activity recognition~\citep{gu2021survey, sztyler2016body}, environmental sensing~\citep{ullo2020advances}, and smart city applications~\citep{syed2021iot}. For example, a scenario that we show in our evaluation is one where data from acoustic and seismic sensors are used to perform {\em joint localization and classification of multiple simultaneous targets\/}, a notoriously hard problem to solve using classical symbolic approaches due to sensory signature overlap. While our representation learning approach is technically independent of the specific sensing modality, we focus on {\em omnidirectional\/} sensors that report time-series data, thereby causing overlapped signatures at each sensor. 

We note that representation learning in distributed sensing systems, such as the above, faces an interesting problem. Namely, while the physical state of the environment is unique, the signals measured by sensors depend on {\em sensor placement\/}. Thus, it is insufficient to only represent sensor signals. Instead, one must map {\em both measured sensor streams and sensor vantage points\/} into the latent space, wherein dependencies between signals and observation locations can be properly represented. We show, in the evaluation, that joint signal and location embedding is preferred to solutions that use locations as mere {\em external metadata\/} associated with sensor streams, as it enables downstream learning-based analytics to jointly reason about locations in the same latent space. Importantly, at encoder training time, the exact deployment-time sensor placement is not known. Thus, the encoder should be trained without access to future deployment-time placement information. This requirement differentiates our approach from popular state-of-the-art solutions (e.g., from the vision literature), such as variants of NeRF~\cite{mildenhall2021nerf} and Gaussian Splatting~\cite{chen2024survey} that are trained to represent an environment only after multi-view observations have already been collected {\em from that particular environment\/}. Our focus on omni-directional sensors, like microphones and seismometers (instead of directional ones like cameras), is another difference from the above work. 

The work can alternatively be construed as advancing {\em positional representation of IoT (sensor) data\/}. In large language models~\cite{zhao2026survey}, text is usually tokenized into a string, where tokens are augmented with one-dimensional positional embeddings. In vision transformers~\cite{khan2022transformers}, images are usually tokenized into patches that comprise a two-dimensional grid augmented with positional embeddings denoting patch positions within the image. In sensing, however, different sensors sample the environment at locations that do not generally have a regular linear or grid structure. Thus, we embed location information explicitly in the latent space to augment signal embeddings, so that downstream tasks can jointly reason about signals and locations. The question is: how best to embed location information? What encoder training objective creates a suitable location embedding usable by downstream tasks? How to keep the representation consistent despite minor variations among sensors?

Addressing these challenges, our self-supervised representation learning framework is pre-trained with (unlabeled) multi-vantage data with an objective driven by {\em the duality between signals and vantage points\/}, treating {\em spatial layout as content\/} (similarly to measured sensor signals). The framework allows sensors to be multimodal. To offer
robustness to certain forms of sensor variability within the same modality (which we henceforth call {\em within-modality heterogeneity\/}), it includes an optional calibration capability that keeps the
produced representation consistent despite some variations among sensor properties. The encoder is pre-trained once and can be used for different subsequent sensor placements without further modification. Downstream analytics can be added as simple task-specific heads on top of the same latent space. Individual heads are fine-tuned (with minimal amounts of labeled data) to their respective tasks (such as classification versus localization) upon deployment. Figure~\ref{fig:motivation} summarizes this workflow.

We evaluate our solution empirically on three real-world datasets and a field study. We show that \MODEL{} consistently outperforms eight baselines across diverse modalities, spatial scales, and tasks, with the largest gains on spatial analytics, the primary target of this work, and no adverse effect on recognition tasks. The field study demonstrates that the overheads of the approach allow it to run in real-time on edge hardware. An information-theoretic interpretation of our design is also presented.

The rest of this paper is organized as follows. Section~\ref{sec:method} summarizes the proposed representation learning algorithm. Section~\ref{sec:experiment} evaluates it in the context of IoT tasks that must analyze properties and/or locations of entities in the environment. Section~\ref{sec:rw} presents related work. Section~\ref{sec:limitations} discusses key takeaways and limitations of our approach. The paper concludes in Section~\ref{sec:conclusion}.

\section{Design}\label{sec:method}
Our design rethinks
self-supervised representation learning to produce a placement-aware representation of spatially-distributed sensor data streams that collectively offer a sparse, irregular multimodal sampling of the physical environment. Its novelty lies in representing location information in the latent space along with sensor signals. Thus, besides the sensory signatures, each sensor is determined by both its {\em spatial location\/} (e.g., GPS coordinates of a seismic
station dictating which parts of the crust it samples) and its {\em configuration\/}, which we call structural calibration information. In order to appropriately embed location information into the latent space, two research questions must be answered: 
\begin{enumerate}
\item
What encoder training objective creates a suitable
location embedding usable by downstream tasks? 
\item
How to keep the representation consistent despite minor variations among same-modality sensors (e.g., small changes in sensitivity, resolution, and other characteristics)?
\end{enumerate}

\noindent
{\em To address the first question\/}, our intuition is that locations must be embedded in a manner that directly helps interpret signals measured at that location. In short, the embedding must be guided by one core principle: {\em the duality between signals and vantage points.\/} Vantage points are not
orthogonal metadata to the signals, but rather stand in an equal and mutually-determining relationship with
signals. This principle is
intrinsically applicable across many distributed sensing applications.
Building on this principle, we introduce a dual reconstruction objective during encoder training that asks for reconstruction of signals from placement information and vice
versa. The approach relies on a variant of masked auto-encoders (MAEs)~\citep{he2022masked}, one of the most widely adopted representation learning paradigms in both mainstream AI and IoT sensing~\citep{huang2022masked, feichtenhofer2022masked, kara2024freqmae, kara2024phymask}. Training randomly masks a subset of tokens; a signal decoder is then given the positions of the masked tokens and asked to reconstruct their signals, while a spatial decoder is given the signals of the masked tokens and asked to reconstruct their positions. Reconstructing the masked data forces the auto-encoder to learn the duality between observation locations and observed signals, which is roughly equivalent
to learning signal propagation characteristics through space. Once that is learned, the encoder can construct
a representation of the environment that allows joint reasoning about locations and observations. The approach works with multimodal signals, where each modality is
first encoded separately by its own per-modality encoder, and the resulting per-modality embeddings are then fused across modalities by a joint encoder into a shared representation. 
{\em To address the second question\/}, calibration tokens are introduced that capture sensor-specific characteristics and act to calibrate the latent representation produced by the encoder in a manner that masks sensor variability.

\begin{figure*}[!tbp]
    \centering
    \includegraphics[width=0.9\linewidth]{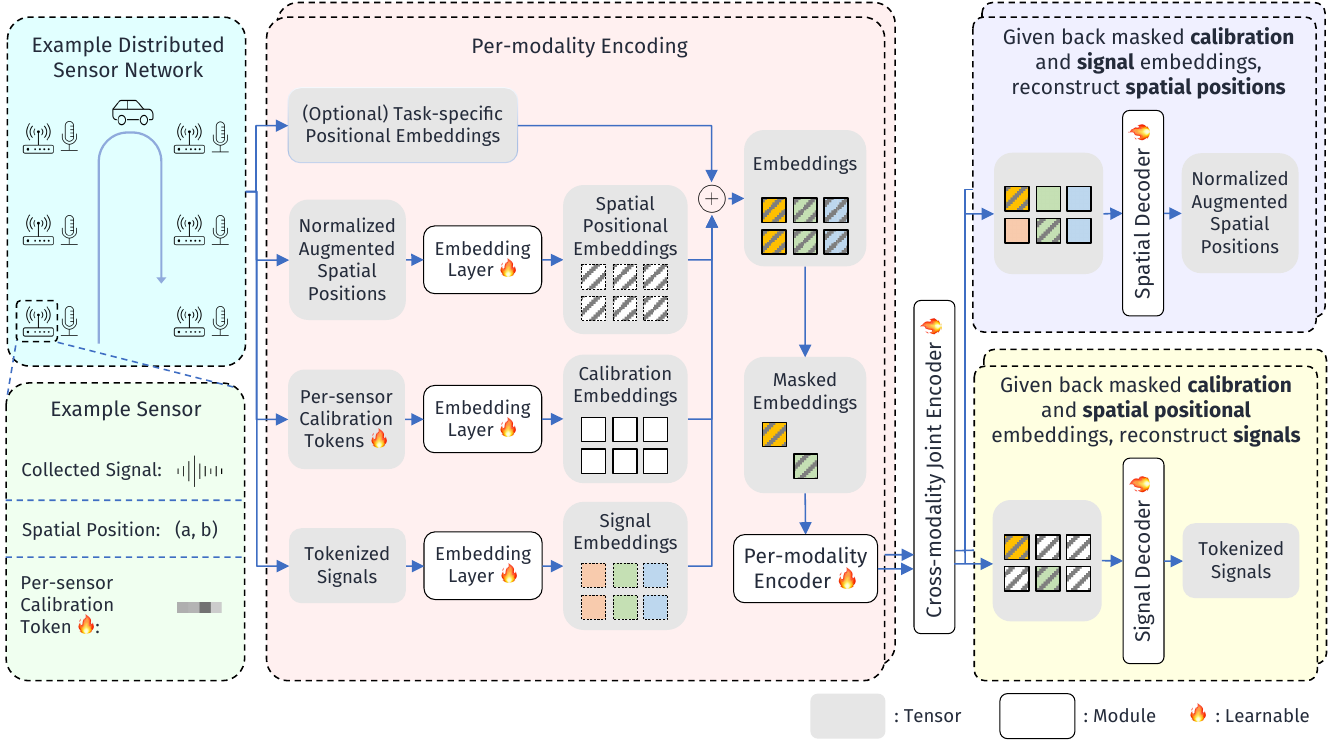}
     \vspace{-0.1in}
    \caption{\small Architectural overview of \MODEL{}. For each modality, tokenized signals, normalized and augmented spatial positions, per-sensor calibration tokens, and any task-specific positional embeddings (e.g., 2D patch positions for spectrograms) are projected through learnable embedding layers, summed, and randomly masked. A per-modality encoder maps the visible tokens to per-modality embeddings, which a joint encoder fuses across modalities into a single \emph{latent representation}. Two decoders reconstruct the masked content from the same representation: the \textbf{signal decoder} consumes it together with the originally-masked spatial-position and calibration embeddings; the \textbf{spatial decoder} consumes it together with the originally-masked signal and calibration embeddings. Because the per-sensor calibration is available to both encoder and decoder, the representation is not pressured to encode it and focuses on signal content and spatial layout instead.}
    \label{fig:framework}
    \vspace{-0.1in}
\end{figure*}

Putting it together, Figure~\ref{fig:framework} shows the \MODEL{} architecture: signals and positions are tokenized, jointly masked, and reconstructed given each other, and a learnable \textit{per-sensor calibration token} conditions both the encoder and the decoders. \MODEL{} is kept deliberately simple, so that empirical gains can be attributed cleanly to the design rather than to auxiliary engineering choices or increased tuning knobs.

\subsection{Embedding Signals, Spatial Positions, and Per-Sensor Calibration Tokens}\label{sec:method_embed}
Below, we use lowercase or uppercase letters for scalars ($k, K$), bold uppercase for matrices ($\mR, \mS$), bold calligraphic for tensors ($\tR, \tS$), and sans-serif uppercase for random tensor variables ($\mathsf{R}, \mathsf{S}$); $\gF$ with subscripts denotes the forward operation of a neural network module. Appendix~3 (Table~\ref{tb:notation}) summarizes the notation.

We consider a multi-modal distributed sensing system with $K$ modalities, the $k$-th modality contains $n^{(k)}$ sensors. Per-sensor signals are tokenized in a task-specific way---IMU time series into temporal segments, acoustic spectrograms into patches---yielding $\tX^{(k)} \in \sR^{n^{(k)} \times m^{(k)} \times d_\tX^{(k)}}$, where $m^{(k)}$ is the number of tokens and $d_\tX^{(k)}$ the token dimension. A learnable linear projection---modality-specific, but shared across all sensors and tokens within a modality---maps each token into the transformer's embedding space at model dimension $d$, yielding signal embeddings $\widetilde{\tX}^{(k)} \in \sR^{n^{(k)} \times m^{(k)} \times d}$.

Each sensor's \textbf{spatial position} in the field is a continuous vector, collected into $\mS^{(k)} \in \sR^{n^{(k)} \times d_\mS}$ (for example, $d_\mS=2$ for longitudinal and lateral coordinates for a planner deployment, or $d_\mS=3$ for 3-dimensional deployments)---unlike the discrete ordinal indices typical of NLP~\citep{vaswani2017attention} and CV~\citep{dosovitskiy2020image}, for which the standard positional embedding strategies~\citep{vaswani2017attention, dosovitskiy2020image, su2024roformer, press2021train} do not apply. We instead embed positions continuously: after broadcasting $\mS^{(k)}$ to $\tS^{(k)} \in \sR^{n^{(k)} \times m^{(k)} \times d_\mS}$ along the token dimension (i.e., $\tS^{(k)}_{i,j,:} = \mS^{(k)}_{i,:}$ for every token $j$), we employ a separate linear layer (modality-specific, shared across sensors and tokens) to project it to a spatial positional embedding $\widetilde{\tS}^{(k)} \in \sR^{n^{(k)} \times m^{(k)} \times d}$, which is added to the signal embeddings.

Two practical issues remain. First, spatial positions vary widely in absolute location and scale across samples or datasets---small parking lots versus large open areas, different cities---so we \textbf{normalize} each sample's positions to zero mean and unit variance. Second, existing datasets cover only a handful of layouts, on which the spatial embeddings can overfit and lose generalization to unseen arrangements; we therefore mitigate this with \textbf{geometric augmentation}---random rotation and translation of the normalized positions during pretraining.

\emph{Within-modality heterogeneity}, introduced earlier, distinguishes nominally identical sensors of the same modality through hardware-level variation (versions, vendors), deployment-level variation (mounting, terrain coupling, surrounding geology, body-part placement), and runtime variation (voltage, temperature, wind, ambient noise). Hand-labeling these factors is impractical at scale, and ignoring them allows the encoder to conflate them with scene content. Instead, we assign each sensor a learnable \textbf{per-sensor calibration token}, $\mR^{(k)} \in \sR^{n^{(k)} \times d_\mR}$, with $d_\mR \ll d$ for parameter efficiency. As with spatial positions, we broadcast $\mR^{(k)}$ to $\tR^{(k)} \in \sR^{n^{(k)} \times m^{(k)} \times d_\mR}$ and apply a linear projection of the same form to obtain the \textbf{per-sensor calibration embedding}, $\widetilde{\tR}^{(k)} \in \sR^{n^{(k)} \times m^{(k)} \times d}$. The embedding is also added to the signal embedding and---critically---supplied as input to \emph{both} the encoder and the decoder, placing the calibration as conditioning in the manner of conditional autoencoders. Sections~\ref{sec:method_objective}--\ref{sec:theory} formalize the resulting asymmetry between content $(X, S)$ and context $R$ and show how it encourages the encoder to calibrate against the heterogeneity while excluding it from the representation.

\subsection{Masked Autoencoding for Signals and Spatial Positions}\label{sec:method_objective}

We sum the signal, spatial, and calibration embeddings (plus any task-specific positional embeddings such as 2D patch positions in a spectrogram, omitted here for clarity) and apply a binary mask $\mM^{(k)} \in \{0,1\}^{n^{(k)} \times m^{(k)}}$, where $\mM^{(k)}_{ij}=1$ marks a visible token and a fraction $1-\rho$ of the tokens, chosen uniformly at random, is kept visible for a mask ratio $\rho$ ($0.75$ by default). A per-modality transformer encoder $\fenc$ maps the visible tokens to a per-modality embedding $\widetilde{\mZ}^{(k)}$---the encoder's latent embedding in standard MAE terminology:
\[
     \widetilde{\mZ}^{(k)} = \fenc(\mask(\widetilde{\tX}^{(k)}+\widetilde{\tS}^{(k)}+\widetilde{\tR}^{(k)}; \mM^{(k)})),
\]
where $\mask(\tA; \mM)$ gathers the token vectors $\tA_{i,j,:}$ with $\mM_{ij}=1$, in row-major order, into a matrix with $m_\mM^{(k)} = \sum_{i,j} \mM^{(k)}_{ij}$ rows, so that $\widetilde{\mZ}^{(k)} \in \sR^{m_\mM^{(k)} \times d}$. A joint transformer encoder $\fjointenc$ then fuses across modalities into a single \emph{latent representation} $\mZ$, shared by all downstream decoders:
\[
     \mZ = \fjointenc(\concat(\widetilde{\mZ}^{(1)},\dots, \widetilde{\mZ}^{(K)})),
\]
where $\concat(\cdot)$ concatenates along the token axis, so $\mZ$ has $\sum_k m_\mM^{(k)}$ rows and $d$ columns. At fine-tuning, the encoders are frozen, and a task-specific prediction head is trained on $\mZ$ under appropriate supervised objectives.

At pretraining, the latent representation $\mZ$ is decoded so the encoders can be trained in a self-supervised manner. Standard MAE uses a single signal decoder, which neither pushes the representation to encode spatial layout nor leaves room for calibration to enter as conditioning. We instead use two decoders---one for signals, one for spatial positions---and each consumes the same $\mZ$. The \textbf{signal decoder} $\fsigdec$ tries to reconstruct the masked signals, conditioning on $\mZ$ and the part of spatial positional embeddings and per-sensor calibration embeddings that were originally \emph{masked} before encoding:
\[
     \widehat{\tX}^{(k)} = \fsigdec(\concat(\mZ, \mask(\widetilde{\tS}^{(k)}+\widetilde{\tR}^{(k)} ; \overline{\mM}^{(k)}))),
\]
where $\overline{\mM}^{(k)} = 1 - \mM^{(k)}$ is the complementary mask, so $\mask(\cdot;\, \overline{\mM}^{(k)})$ outputs the exact embeddings that $\mask(\cdot;\, \mM^{(k)})$ leaves out; $\widehat{\tX}^{(k)}$ denotes the reconstructed signals. 

Symmetrically, the \textbf{spatial decoder} $\fspdec$ reconstructs the masked spatial positions from $\mZ$, with the originally \emph{masked} part of signal embeddings and the per-sensor calibration embeddings as additional conditions:
\[
     \widehat{\tS}^{(k)} = \fspdec(\concat(\mZ, \mask(\widetilde{\tX}^{(k)} +\widetilde{\tR}^{(k)}; \overline{\mM}^{(k)}))),
\]
where $\widehat{\tS}^{(k)}$ denotes the reconstructed spatial positions. 

The training loss $L$ sums the MSE reconstruction losses of both decoders:
\[
\begin{aligned}
    L = \sum_{k=1}^K & \| \mask(\tX^{(k)}-\widehat{\tX}^{(k)};\overline{\mM}^{(k)}\odot\mO^{(k)})\|_2^2 \\
    + & \| \mask(\tS^{(k)}-\widehat{\tS}^{(k)};\overline{\mM}^{(k)}\odot\mO^{(k)})\|_2^2,
\end{aligned}
\]
where $\mO^{(k)} \in \{0,1\}^{n^{(k)} \times m^{(k)}}$ marks the observed (non-missing) tokens and $\odot$ is the entrywise product, so that missing entries are excluded from the loss.
This \emph{mutual-reconstruction objective} pushes the encoder to understand and preserve both signal and spatial-position information in the latent representation as content. The calibration token plays a complementary role. Providing it to both the encoder and the decoders allows the encoder to use it in interpreting each sensor's signal---calibrating against the sensor's idiosyncrasies---while relieving the encoder of any need to retain the calibration information in the latent representation once this interpretation is complete; we grant the encoder \emph{permission to drop} the calibration. We expect this \emph{permission to drop} to translate into an \emph{action to drop}, since deep networks under standard stochastic training tend not to preserve information that is not needed downstream. This parsimony tendency is empirically supported and exploited in conditional generative models, where feeding a variable---class labels, image attributes, or group identity---to the decoder reliably removes that variable from the encoder's output~\citep{sohn2015cvae, lample2017fader, bouchacourt2018mlvae}. It has also been studied as a general property of deep networks~\citep{shwartzziv2017opening}, and is supported by an information-bottleneck argument on weights~\citep{achille2018emergence}. \MODEL{} turns this tendency to its advantage: rather than memorizing the particular sensor-and-scene combinations observed during pretraining---which would tend to overfit each sensor's idiosyncrasies at the expense of generalization---the latent representation concentrates on signal and spatial-position content and becomes robust to within-modality heterogeneity.

Another practical concern is that missing data due to hardware failures or unreliable links is common in multi-sensor deployments. We address this by padding missing entries with zeros and excluding them from the reconstruction loss through $\mO^{(k)}$. Algorithms~\ref{alg:pretrain} and~\ref{alg:finetune} consolidate the full pretraining and fine-tuning procedures introduced above. Pretraining covers every operational detail (normalization, augmentation, per-modality and joint encoding, mutual reconstruction, missing-data handling); fine-tuning is presented at the abstraction level of a generic head $H$, which maps $\mZ$ to a prediction $\hat{y}$ of the label $y$, and a task-specific loss $\mathcal{L}_{\text{task}}(\hat{y}, y)$, so the same recipe applies across the recognition and localization tasks.

\begin{algorithm}[!htbp]
\small
\caption{- \MODEL{} pretraining procedure.}
\label{alg:pretrain}
\begin{algorithmic}[1]
\Require Unlabeled multi-modal dataset $\{(\tX^{(k)}, \mS^{(k)})\}_{k=1}^{K}$, with observation masks $\{\mO^{(k)}\}_{k=1}^{K}$ marking the observed tokens; mask ratio $\rho$.
\State Initialize per-modality encoders $\fencnk$, joint encoder $\fjointenc$, signal decoders $\fsigdecnk$, spatial decoders $\fspdecnk$, per-modality linear embedding layers (for signals, spatial positions, and calibration tokens), and per-sensor calibration tokens $\{\mR^{(k)}\}_{k=1}^{K}$.
\For{each training step on a mini-batch of $\{(\tX^{(k)}, \mS^{(k)})\}_{k=1}^{K}$}
  \For{$k = 1, \dots, K$}
    \State Pad missing entries in $\tX^{(k)}$ and $\mS^{(k)}$ with zeros.
    \State $\mS^{(k)} \gets \text{normalize}(\mS^{(k)})$
    \State $\mS^{(k)} \gets \text{aug}(\mS^{(k)})$
    \State Linearly embed signals, broadcasted positions, and broadcasted calibration tokens into $\widetilde{\tX}^{(k)},\,\widetilde{\tS}^{(k)},\,\widetilde{\tR}^{(k)}$.
    \State Sample $\mM^{(k)}$ uniformly at random with a fraction $1-\rho$ of its entries set to $1$.
    \State $\widetilde{\mZ}^{(k)} \gets \fenc(\mask(\widetilde{\tX}^{(k)} + \widetilde{\tS}^{(k)} + \widetilde{\tR}^{(k)};\, \mM^{(k)}))$
  \EndFor
  \State $\mZ \gets \fjointenc(\concat(\widetilde{\mZ}^{(1)}, \dots, \widetilde{\mZ}^{(K)}))$
  \For{$k = 1, \dots, K$}
    \State $\widehat{\tX}^{(k)} \gets \fsigdec(\concat(\mZ,\, \mask(\widetilde{\tS}^{(k)}+\widetilde{\tR}^{(k)};\, \overline{\mM}^{(k)})))$
    \State $\widehat{\tS}^{(k)} \gets \fspdec(\concat(\mZ,\, \mask(\widetilde{\tX}^{(k)}+\widetilde{\tR}^{(k)};\, \overline{\mM}^{(k)})))$
  \EndFor
  \State $L_{X} \gets \sum_{k=1}^{K} \big\| \mask(\tX^{(k)}-\widehat{\tX}^{(k)};\, \overline{\mM}^{(k)}\odot\mO^{(k)}) \big\|_2^2$
  \State $L_{S} \gets \sum_{k=1}^{K} \big\| \mask(\tS^{(k)}-\widehat{\tS}^{(k)};\, \overline{\mM}^{(k)}\odot\mO^{(k)}) \big\|_2^2$
  \State $L \gets L_{X} + L_{S}$
  \State Update all encoders, decoders, embedding layers, and $\{\mR^{(k)}\}_{k=1}^{K}$ by gradient descent on $L$.
\EndFor
\State \Return pretrained encoders $\fencnk,\,\fjointenc$, embedding layers, and per-sensor calibration tokens.
\end{algorithmic}

\end{algorithm}
\vspace{-0.15in}

\begin{algorithm}[!htbp]
\small
\caption{- \MODEL{} fine-tuning procedure for an arbitrary downstream task.}
\label{alg:finetune}
\begin{algorithmic}[1]
\Require Labeled dataset $\{(\{(\tX^{(k)}, \mS^{(k)})\}_{k=1}^{K},\, y)\}$; pretrained encoders $\fencnk,\,\fjointenc$, embedding layers, and per-sensor calibration tokens $\{\mR^{(k)}\}_{k=1}^{K}$; task-specific loss $\mathcal{L}_{\text{task}}$.
\State Freeze all parameters obtained from pretraining.
\State Initialize a task-specific prediction head $H$.
\For{each fine-tuning step on a mini-batch}
  \For{$k = 1, \dots, K$}
    \State Pad missing entries in $\tX^{(k)}$ and $\mS^{(k)}$ with zeros.
    \State $\mS^{(k)} \gets \text{normalize}(\mS^{(k)})$
    \State Linearly embed signals, broadcasted positions, and broadcasted calibration tokens into $\widetilde{\tX}^{(k)},\,\widetilde{\tS}^{(k)},\,\widetilde{\tR}^{(k)}$.
    \State $\widetilde{\mZ}^{(k)} \gets \fenc(\widetilde{\tX}^{(k)} + \widetilde{\tS}^{(k)} + \widetilde{\tR}^{(k)})$
  \EndFor
  \State $\mZ \gets \fjointenc(\concat(\widetilde{\mZ}^{(1)}, \dots, \widetilde{\mZ}^{(K)}))$
  \State $\hat{y} \gets H(\mZ)$
  \State Update $H$ only, by gradient descent on $\mathcal{L}_{\text{task}}(\hat{y}, y)$.
\EndFor
\State \Return fine-tuned task head $H$.
\end{algorithmic}

\end{algorithm}
\vspace{-0.15in}

\begin{table*}[!t]
\caption{\small Per-dataset specifications.}
\vspace{-0.1in}
\label{tb:datasets}
\begin{center}
\begin{adjustbox}{max width=\linewidth}
\begin{tabular}{lccc}
\toprule
 & M3N-VC~\citep{li2025restoreml} & Ridgecrest~\citep{caltech1926socal} & RealWorld-HAR~\citep{sztyler2016body} \\
\midrule
Spatial scale       & $30$--$80$\,m-radius & $80$\,km-radius & sub-meter \\
Modalities          & microphone $+$ geophone & broadband seismometer $+$ accelerometer & accelerometer $+$ gyroscope $+$ magnetometer \\
Sampling rate       & $16$\,kHz / $200$\,Hz & $100$\,Hz / $100$\,Hz & $50$\,Hz / $50$\,Hz / $50$\,Hz \\
Nodes per sample    & $6$--$8$ & $16$ & $6$\\
Sample length       & $2$\,s & $30.72$\,s & $4$\,s \\
Total samples       & $21{,}694$ & $31{,}452$ & $13{,}351$ \\
Train / val split   & temporal ($\approx$ $3{:}1$) & temporal ($\approx$ $7{:}3$) & subject-based ($10{:}5$) \\
Downstream tasks    & (single-/multi-vehicle) recognition and localization  & epicenter localization & activity recognition \\
\bottomrule
\end{tabular}
\end{adjustbox}
\end{center}
\vspace{-0.1in}
\end{table*}

\subsection{Theoretical Analysis}\label{sec:theory}
We analyze \MODEL{} in comparison to classical MAE through the lens of information theory, as formalized in the following proposition:
\begin{proposition} \label{pp:information}
Let $\mathsf{X}^{(k)}, \mathsf{S}^{(k)}, \mathsf{R}^{(k)}$ denote the random variables corresponding to the signals, the spatial positions, and the per-sensor calibration tokens, for $k \in \{1, \dots, K\}$, respectively. Let $\mathsf{Z}$ denote the random variable corresponding to the latent representation $\mZ$ in either classical MAE or \MODEL{}.
Let $\E[L']$ and $\E[L]$ denote the expected losses of classical MAE and \MODEL{} over the data distribution, respectively, and let $C'$ and $C$ be constants independent of model parameters. Then, under certain assumptions, for classical MAE, we can have the following bound:
\[
    -\E[L'] + C' \leq \sum_{k=1}^K I(\mathsf{X}^{(k)}; \mathsf{Z}),
\]
where $I(\cdot;\cdot)$ denotes mutual information. In contrast, for \MODEL{}, we can have
\[
\begin{aligned}
    -\E[L] + C \leq \sum_{k=1}^K & I(\mathsf{X}^{(k)}; \mathsf{Z} \mid \mathsf{S}^{(k)}, \mathsf{R}^{(k)}) \\
    + & I(\mathsf{S}^{(k)}; \mathsf{Z} \mid \mathsf{X}^{(k)}, \mathsf{R}^{(k)}).
\end{aligned}
\]
where $I(\cdot;\cdot|\cdot)$ denotes conditional mutual information.
\end{proposition}

The proof is detailed in Appendix~1. The two bounds differ structurally in a way that maps cleanly to \MODEL{}'s design. In classical MAE, the bound rewards the representation for retaining information about the input signals, with no further structure. In \MODEL{}, every mutual-information term in the bound \emph{conditions on the calibration tokens} $\mathsf{R}^{(k)}$, so the bound rewards the representation for capturing signal and spatial-position information \emph{beyond} what calibration $\mathsf{R}^{(k)}$ already explains. The architectural asymmetry between $(X,S)$ and $R$ is precisely what the design principle asks for. Together, the two mutual information terms encourage embeddings that are jointly informative of signals and spatial layout while sparing the representation from memorizing what the calibration tokens already supply, hence encouraging a representation that is {\em robust\/} to sensor heterogeneity. The bound does not by itself show that calibration information is absent from the representation; that expectation rests on the parsimony argument of Section~\ref{sec:method_objective} and the experiments of Section~\ref{exp:ablation}.

\section{Evaluation}\label{sec:experiment}
The evaluation asks five fundamental questions: (1) Does the proposed representation help improve the accuracy of downstream analytics? (2) To what degree is encoder pre-training really independent of future sensor placement? (3) To what degree do the different components of the scheme contribute to its performance advantages (i.e., an ablation study)? (4) What is the run-time overhead of the proposed scheme? (5) Do the results generalize across applications? Below, we detail the experimental setup, then proceed with answering the above questions in subsequent sections, respectively. 

\subsection{Experimental Setup}\label{exp:setup}

\noindent\textbf{Datasets.} We evaluate \MODEL{} on three real-world multi-modal distributed sensing datasets that span orders of magnitude in spatial scale and exercise complementary modality stacks; Table~\ref{tb:datasets} summarizes their specifications. M3N-VC~\citep{li2025restoreml} pairs acoustic and vibration recordings from $6$--$8$ nodes deployed across six outdoor sites, with 1\,Hz vehicle GPS tracks supplying meter-level labels for four-vehicle-plus-background recognition and several spatial-prediction variants. The Ridgecrest Seismicity Dataset~\citep{caltech1926socal} pairs broadband seismometer and accelerometer waveforms from 16 stations across an $80$\,km-radius region for kilometer-scale earthquake localization, with a year-based temporal split that guards against event leakage. RealWorld-HAR~\citep{sztyler2016body} provides accelerometer, gyroscope, and magnetometer signals from six body-mounted nodes on 15 subjects performing eight activities; a subject-based split (10 / 5) tests generalization to unseen people.

\smallskip
\noindent\textbf{Baselines.} We compare \MODEL{} against eight self-supervised representation learning baselines covering multiple paradigms and domains as follows:
\begin{itemize}[leftmargin=*]
\item
\textbf{CMC}~\citep{tian2020contrastive}: Learns shared representations by maximizing mutual information between views, enabling view-agnostic and scalable contrastive learning across multiple modalities.
\item
\textbf{Cosmo}~\citep{ouyang2022cosmo}: Integrates contrastive feature alignment with attention-based selective fusion to effectively capture shared and distinctive patterns from multimodal data under scarce labeling.
\item
\textbf{SimCLR}~\citep{chen2020simple}: Forms discriminative visual embeddings by aligning augmented image pairs through a nonlinear projection and optimizing the NT-Xent contrastive loss.
\item
\textbf{AudioMAE}~\citep{huang2022masked}: Applies masked autoencoding to audio by operating on spectrogram patches, using a Transformer to reconstruct masked regions and capture time-frequency patterns without relying on external modalities.
\item
\textbf{CAV-MAE}~\citep{gong2022contrastive}: Combines masked autoencoding and contrastive learning in a unified audio-visual framework, using modality-specific encoders and a joint decoder to learn both fused and aligned representations from spectrogram and image patches.
\item
\textbf{FOCAL}~\citep{liu2023focal}: Separates multimodal time-series signals into shared and private latent spaces, enforcing orthogonality and applying contrastive and temporal constraints to capture both modality-consistent and modality-exclusive features.
\item
\textbf{FreqMAE}~\citep{kara2024freqmae}: Enhances MAEs for multimodal sensing by incorporating frequency-aware transformers, factorized fusion of shared and private features, and a weighted loss that prioritizes informative frequency bands and high-SNR samples.
\item
\textbf{PhyMask}~\citep{kara2024phymask}: Improves masked autoencoding by adaptively selecting time-frequency patches based on energy and coherence metrics, enabling efficient and informative masking tailored to physical sensing signals.
\end{itemize}
The above list was curated to include both contrastive methods (CMC~\citep{tian2020contrastive}, Cosmo~\citep{ouyang2022cosmo}, SimCLR~\citep{chen2020simple}, FOCAL~\citep{liu2023focal}) and masked-autoencoders (AudioMAE~\citep{huang2022masked}, CAV-MAE~\citep{gong2022contrastive}, FreqMAE~\citep{kara2024freqmae}, PhyMask~\citep{kara2024phymask}). It spans both general-purpose time-series and multimodal representation learning approaches (AudioMAE, CAV-MAE, CMC, SimCLR) and approaches tailored for IoT sensing (Cosmo, FOCAL, FreqMAE, PhyMask). 

\smallskip
\noindent\textbf{Backbone and Task Heads.} All methods, including \MODEL{}, share a standard ViT backbone~\citep{dosovitskiy2020image} and identical task-specific prediction heads. These heads are kept deliberately lightweight (a linear classifier or a small transformer) so that performance differences reflect the pretrained representations rather than head capacity.

\smallskip
\noindent\textbf{Implementation.} We pretrain and fine-tune every method for the same sufficient number of epochs and aggregate all reported numbers over three random seeds.

\subsection{Impact on Downstream Accuracy}\label{exp:m3n-vc}
To understand how our placement-aware latent representation impacts the accuracy of different downstream tasks, in this section, we opted for experimentation with the Multi-Modal Multi-Node Vehicle Classification (M3N-VC) dataset~\citep{li2025restoreml}, due to its support for both localization and classification problems, as well as multiple targets. The dataset comprises acoustic and seismic data on passing vehicles, collected from March 2023 to October 2024 across six outdoor sites spanning varied terrain (asphalt, dirt, gravel, concrete) and weather (sunny, rainy, windy). In this dataset, per-sensor recordings are segmented into non-overlapping $2$-second clips and transformed into spectrograms for model input. Per-node GPS coordinates are converted into meter-level local positions via Local Tangent Plane projection~\citep{united1987department}. 
Unless otherwise noted, the encoder is pretrained on data from all six sites. A task-specific head is then fine-tuned for each respective task. Three tasks are tested of increasing complexity, namely: (i) single-vehicle localization, (ii) single-vehicle classification, and (iii) joint multi-vehicle classification and localization, as described below. Spatial analytics are the primary target of the design; the classification task is included to verify that the design does not compromise tasks with little spatial dependence.

\definecolor{lightblue}{HTML}{E6F0FF}
\newcommand{\mygray}[1]{\textcolor{gray}{\footnotesize #1}}

\begin{table*}[!tb]
\caption{\small Single-vehicle localization on M3N-VC under shrinking supervision (label ratio $1.0 \to 0.2$).}
\vspace{-0.1in}
\label{tb:m3n-vc1}
\begin{center}
\begin{adjustbox}{max width=0.8\linewidth}
\begin{tabular}{lcccccc}
\toprule
\multirow{3}{*}{Method} & \multicolumn{6}{c}{M3N-VC Single-vehicle Localization} \\ \cmidrule(lr){2-7}
                        & \multicolumn{2}{c}{Label Ratio 1.0} & \multicolumn{2}{c}{Label Ratio 0.5} & \multicolumn{2}{c}{Label Ratio 0.2} \\ \cmidrule(lr){2-3} \cmidrule(lr){4-5} \cmidrule(lr){6-7}
                        & \multicolumn{1}{c}{MSE ($m^2$) ($\downarrow$)} & \multicolumn{1}{c}{Dist. Err. ($m$) ($\downarrow$)} & \multicolumn{1}{c}{MSE ($m^2$) ($\downarrow$)} & \multicolumn{1}{c}{Dist. Err. ($m$) ($\downarrow$)} & \multicolumn{1}{c}{MSE ($m^2$) ($\downarrow$)} & \multicolumn{1}{c}{Dist. Err. ($m$) ($\downarrow$)} \\
\midrule
CMC                     & \multicolumn{1}{c}{51.11 \mygray{$\pm$ 14.67}} & \multicolumn{1}{c}{6.76 \mygray{$\pm$ 0.75}} & \multicolumn{1}{c}{71.81 \mygray{$\pm$ 15.32}} & \multicolumn{1}{c}{7.99 \mygray{$\pm$ 0.64}} & \multicolumn{1}{c}{111.37 \mygray{$\pm$ 8.02}} & \multicolumn{1}{c}{11.05 \mygray{$\pm$ 0.57}} \\
Cosmo                   & \multicolumn{1}{c}{38.40 \mygray{$\pm$ 4.14}} & \multicolumn{1}{c}{6.03 \mygray{$\pm$ 0.21}} & \multicolumn{1}{c}{53.12 \mygray{$\pm$ 9.75}} & \multicolumn{1}{c}{7.19 \mygray{$\pm$ 0.40}} & \multicolumn{1}{c}{97.08 \mygray{$\pm$ 9.49}} & \multicolumn{1}{c}{10.95 \mygray{$\pm$ 0.57}} \\
SimCLR                  & \multicolumn{1}{c}{34.40 \mygray{$\pm$ 4.47}} & \multicolumn{1}{c}{5.64 \mygray{$\pm$ 0.25}} & \multicolumn{1}{c}{45.14 \mygray{$\pm$ 7.34}} & \multicolumn{1}{c}{6.57 \mygray{$\pm$ 0.08}} & \multicolumn{1}{c}{74.53 \mygray{$\pm$ 3.13}} & \multicolumn{1}{c}{9.48 \mygray{$\pm$ 0.17}} \\
AudioMAE                & \multicolumn{1}{c}{22.36 \mygray{$\pm$ 0.49}} & \multicolumn{1}{c}{5.40 \mygray{$\pm$ 0.11}} & \multicolumn{1}{c}{30.12 \mygray{$\pm$ 2.97}} & \multicolumn{1}{c}{6.33 \mygray{$\pm$ 0.28}} & \multicolumn{1}{c}{41.75 \mygray{$\pm$ 3.30}} & \multicolumn{1}{c}{7.47 \mygray{$\pm$ 0.28}} \\
CAV-MAE                 & \multicolumn{1}{c}{18.85 \mygray{$\pm$ 0.41}} & \multicolumn{1}{c}{5.06 \mygray{$\pm$ 0.04}} & \multicolumn{1}{c}{22.90 \mygray{$\pm$ 0.82}} & \multicolumn{1}{c}{5.58 \mygray{$\pm$ 0.12}} & \multicolumn{1}{c}{24.84 \mygray{$\pm$ 0.33}} & \multicolumn{1}{c}{5.78 \mygray{$\pm$ 0.10}} \\
FOCAL                   & \multicolumn{1}{c}{32.43 \mygray{$\pm$ 4.68}} & \multicolumn{1}{c}{5.37 \mygray{$\pm$ 0.22}} & \multicolumn{1}{c}{40.84 \mygray{$\pm$ 2.82}} & \multicolumn{1}{c}{6.20 \mygray{$\pm$ 0.19}} & \multicolumn{1}{c}{69.62 \mygray{$\pm$ 5.62}} & \multicolumn{1}{c}{8.50 \mygray{$\pm$ 0.35}} \\
FreqMAE                 & \multicolumn{1}{c}{29.61 \mygray{$\pm$ 2.85}} & \multicolumn{1}{c}{5.36 \mygray{$\pm$ 0.16}} & \multicolumn{1}{c}{42.06 \mygray{$\pm$ 14.44}} & \multicolumn{1}{c}{6.25 \mygray{$\pm$ 0.70}} & \multicolumn{1}{c}{91.40 \mygray{$\pm$ 35.32}} & \multicolumn{1}{c}{9.15 \mygray{$\pm$ 1.27}} \\
PhyMask                 & \multicolumn{1}{c}{28.02 \mygray{$\pm$ 5.91}} & \multicolumn{1}{c}{5.29 \mygray{$\pm$ 0.33}} & \multicolumn{1}{c}{33.74 \mygray{$\pm$ 2.18}} & \multicolumn{1}{c}{5.85 \mygray{$\pm$ 0.12}} & \multicolumn{1}{c}{64.36 \mygray{$\pm$ 4.70}} & \multicolumn{1}{c}{8.44 \mygray{$\pm$ 0.36}} \\
\rowcolor{lightblue}
\MODEL                  & \multicolumn{1}{c}{\textbf{12.98 \mygray{$\pm$ 0.11}}} & \multicolumn{1}{c}{\textbf{4.20 \mygray{$\pm$ 0.07}}} & \multicolumn{1}{c}{\textbf{15.07 \mygray{$\pm$ 1.03}}} & \multicolumn{1}{c}{\textbf{4.51 \mygray{$\pm$ 0.09}}} & \multicolumn{1}{c}{\textbf{21.36 \mygray{$\pm$ 0.62}}} & \multicolumn{1}{c}{\textbf{5.40 \mygray{$\pm$ 0.04}}} \\
\bottomrule
\end{tabular}
\end{adjustbox}
\end{center}
\vspace{-0.1in}
\end{table*}

\begin{table}[!tb]
\caption{\small Multi-vehicle joint recognition and localization on M3N-VC; mAP@\emph{r} (\%) ($\uparrow$) with spatial tolerance $r \in \{4,6,8,10\}$~m.}
\vspace{-0.1in}
\label{tb:m3n-vc2}
\begin{center}
\begin{adjustbox}{max width=\linewidth}
\begin{tabular}{lcccc}
\toprule
\multirow{2}{*}{Method} & \multicolumn{4}{c}{M3N-VC Multi-vehicle Joint Recognition and Localization}                                                                                                                                      \\ \cmidrule(lr){2-5}
                        & \multicolumn{1}{c}{mAP@4m  } & \multicolumn{1}{c}{mAP@6m } & \multicolumn{1}{c}{mAP@8m } & \multicolumn{1}{c}{mAP@10m  } \\
\midrule
CMC                     & \multicolumn{1}{c}{0.06 \mygray{$\pm$ 0.05}} & \multicolumn{1}{c}{0.48 \mygray{$\pm$ 0.36}} & \multicolumn{1}{c}{1.61 \mygray{$\pm$ 1.10}} & \multicolumn{1}{c}{3.62 \mygray{$\pm$ 2.19}} \\
Cosmo                   & \multicolumn{1}{c}{0.16 \mygray{$\pm$ 0.05}} & \multicolumn{1}{c}{1.66 \mygray{$\pm$ 0.23}} & \multicolumn{1}{c}{4.77 \mygray{$\pm$ 0.72}} & \multicolumn{1}{c}{9.52 \mygray{$\pm$ 1.20}} \\
SimCLR                  & \multicolumn{1}{c}{0.31 \mygray{$\pm$ 0.14}} & \multicolumn{1}{c}{2.22 \mygray{$\pm$ 0.58}} & \multicolumn{1}{c}{6.53 \mygray{$\pm$ 1.24}} & \multicolumn{1}{c}{13.07 \mygray{$\pm$ 2.08}} \\
AudioMAE                & \multicolumn{1}{c}{1.39 \mygray{$\pm$ 0.48}} & \multicolumn{1}{c}{6.96 \mygray{$\pm$ 1.42}} & \multicolumn{1}{c}{17.11 \mygray{$\pm$ 3.24}} & \multicolumn{1}{c}{28.98 \mygray{$\pm$ 4.01}} \\
CAV-MAE                 & \multicolumn{1}{c}{22.12 \mygray{$\pm$ 2.94}} & \multicolumn{1}{c}{52.08 \mygray{$\pm$ 4.16}} & \multicolumn{1}{c}{73.41 \mygray{$\pm$ 3.24}} & \multicolumn{1}{c}{85.36 \mygray{$\pm$ 1.78}} \\
FOCAL                   & \multicolumn{1}{c}{0.08 \mygray{$\pm$ 0.05}} & \multicolumn{1}{c}{0.82 \mygray{$\pm$ 0.40}} & \multicolumn{1}{c}{2.94 \mygray{$\pm$ 1.04}} & \multicolumn{1}{c}{6.82 \mygray{$\pm$ 1.99}} \\
FreqMAE                 & \multicolumn{1}{c}{0.24 \mygray{$\pm$ 0.01}} & \multicolumn{1}{c}{1.67 \mygray{$\pm$ 0.32}} & \multicolumn{1}{c}{5.34 \mygray{$\pm$ 0.99}} & \multicolumn{1}{c}{11.31 \mygray{$\pm$ 1.49}} \\
PhyMask                 & \multicolumn{1}{c}{0.08 \mygray{$\pm$ 0.03}} & \multicolumn{1}{c}{0.88 \mygray{$\pm$ 0.24}} & \multicolumn{1}{c}{3.04 \mygray{$\pm$ 0.74}} & \multicolumn{1}{c}{6.64 \mygray{$\pm$ 1.46}} \\ \rowcolor{lightblue}
\MODEL                  & \multicolumn{1}{c}{\textbf{41.57 \mygray{$\pm$ 2.69}}} & \multicolumn{1}{c}{\textbf{71.82 \mygray{$\pm$ 3.69}}} & \multicolumn{1}{c}{\textbf{86.28 \mygray{$\pm$ 1.77}}} & \multicolumn{1}{c}{\textbf{92.99 \mygray{$\pm$ 0.79}}} \\
\bottomrule
\end{tabular}
\end{adjustbox}
\end{center}
\vspace{-0.1in}
\end{table}

\begin{table}[!htb]
\caption{\small Generalization to a sensor placement not observed during pretraining (M3N-VC sites H08 and H24 held out from pretraining; fine-tuned and evaluated on H24).}
\vspace{-0.1in}
\label{tb:m3n-vc5}
\begin{center}
\begin{adjustbox}{max width=0.8\linewidth}
\begin{tabular}{lcc}
\toprule
\multirow{3}{*}{Method} & \multicolumn{2}{c}{M3N-VC Single-vehicle Localization} \\
                        & \multicolumn{2}{c}{(Finetuned and Evaluated on Unseen Placement)} \\ \cmidrule(lr){2-3}
                        & \multicolumn{1}{c}{MSE ($m^2$) ($\downarrow$)} & \multicolumn{1}{c}{Dist. Err. ($m$) ($\downarrow$)} \\
\midrule
CMC                     & \multicolumn{1}{c}{61.78 \mygray{$\pm$ 17.68}} & \multicolumn{1}{c}{7.23 \mygray{$\pm$ 0.78}} \\
Cosmo                   & \multicolumn{1}{c}{63.43 \mygray{$\pm$ 12.85}} & \multicolumn{1}{c}{7.22 \mygray{$\pm$ 0.63}} \\
SimCLR                  & \multicolumn{1}{c}{35.82 \mygray{$\pm$ 7.57}} & \multicolumn{1}{c}{5.92 \mygray{$\pm$ 0.39}} \\
AudioMAE                & \multicolumn{1}{c}{41.25 \mygray{$\pm$ 4.60}} & \multicolumn{1}{c}{6.64 \mygray{$\pm$ 0.31}} \\
CAV-MAE                 & \multicolumn{1}{c}{37.01 \mygray{$\pm$ 0.68}} & \multicolumn{1}{c}{6.27 \mygray{$\pm$ 0.04}} \\
FOCAL                   & \multicolumn{1}{c}{41.79 \mygray{$\pm$ 11.04}} & \multicolumn{1}{c}{5.91 \mygray{$\pm$ 0.41}} \\
FreqMAE                 & \multicolumn{1}{c}{30.65 \mygray{$\pm$ 1.14}} & \multicolumn{1}{c}{5.51 \mygray{$\pm$ 0.19}} \\
PhyMask                 & \multicolumn{1}{c}{34.83 \mygray{$\pm$ 8.81}} & \multicolumn{1}{c}{5.60 \mygray{$\pm$ 0.38}} \\ \rowcolor{lightblue}
\MODEL                  & \multicolumn{1}{c}{\textbf{21.76 \mygray{$\pm$ 1.00}}} & \multicolumn{1}{c}{\textbf{5.09 \mygray{$\pm$ 0.10}}} \\
\bottomrule
\end{tabular}
\end{adjustbox}
\end{center}
\vspace{-0.1in}
\end{table}

\begin{table}[!htb]
\caption{\small Ablation on M3N-VC single-vehicle localization.}
\vspace{-0.1in}
\label{tb:m3n-vc6}
\begin{center}
\begin{adjustbox}{max width=0.9\linewidth}
\begin{tabular}{lcc}
\toprule
\multirow{2}{*}{Method} & \multicolumn{2}{c}{Single-vehicle Localization}                                                                \\ \cmidrule(lr){2-3}
                        & \multicolumn{1}{c}{MSE  ($\downarrow$)} & \multicolumn{1}{c}{Dist. Err.  ($\downarrow$)} \\
\midrule
\MODEL                                      & \multicolumn{1}{c}{12.98 \mygray{$\pm$ 0.11}} & \multicolumn{1}{c}{4.20 \mygray{$\pm$ 0.07}} \\ \midrule
\textit{w/o} Geometric Augmentation in Pretrain      & \multicolumn{1}{c}{15.59 \mygray{$\pm$ 0.56}} & \multicolumn{1}{c}{4.67 \mygray{$\pm$ 0.04}} \\
\textit{w/o} Reconstructing Spatial Positions        & \multicolumn{1}{c}{14.73 \mygray{$\pm$ 0.35}} & \multicolumn{1}{c}{4.62 \mygray{$\pm$ 0.03}} \\
\textit{w/o} Spatial Positional Embedding            & \multicolumn{1}{c}{15.12 \mygray{$\pm$ 0.58}} & \multicolumn{1}{c}{4.67 \mygray{$\pm$ 0.07}} \\
\textit{w/o} Per-Sensor Calibration Embedding          & \multicolumn{1}{c}{22.55 \mygray{$\pm$ 2.98}} & \multicolumn{1}{c}{5.08 \mygray{$\pm$ 0.13}} \\
+ Sensor-Drop Masking                         & \multicolumn{1}{c}{17.71 \mygray{$\pm$ 3.17}} & \multicolumn{1}{c}{4.82 \mygray{$\pm$ 0.33}} \\
+ Sensor-Balanced Masking                     & \multicolumn{1}{c}{\textbf{12.54 \mygray{$\pm$ 1.63}}} & \multicolumn{1}{c}{\textbf{4.14 \mygray{$\pm$ 0.21}}} \\
+ Mask Ratio 0.85                           & \multicolumn{1}{c}{13.89 \mygray{$\pm$ 2.17}} & \multicolumn{1}{c}{4.35 \mygray{$\pm$ 0.24}} \\
+ Mask Ratio 0.5                            & \multicolumn{1}{c}{14.81 \mygray{$\pm$ 2.57}} & \multicolumn{1}{c}{4.45 \mygray{$\pm$ 0.31}} \\
\bottomrule
\end{tabular}
\end{adjustbox}
\end{center}
\vspace{-0.1in}
\end{table}

\begin{table*}[!t]
\caption{\small Cross-dataset comparison against eight self-supervised pretraining baselines on (1) M3N-VC single-vehicle recognition, (2) Ridgecrest earthquake localization, and (3) RealWorld-HAR activity recognition. \MODEL{} achieves the best or comparable performance across tasks and metrics.}
\vspace{-0.1in}
\label{tb:all}
\begin{center}
\begin{adjustbox}{max width=0.8\linewidth}
\begin{tabular}{lcccccc}
\toprule
\multirow{2}{*}{Method} 
& \multicolumn{2}{c}{M3N-VC Recognition}
& \multicolumn{2}{c}{Ridgecrest Earthquake Localization} 
& \multicolumn{2}{c}{RealWorld-HAR Recognition} \\ 
\cmidrule(lr){2-3} \cmidrule(lr){4-5} \cmidrule(lr){6-7}
& Accuracy (\%) ($\uparrow$) & F1 (\%) ($\uparrow$) 
& MSE ($km^2$) ($\downarrow$) & Dist. Err. ($km$) ($\downarrow$) 
& Accuracy (\%) ($\uparrow$) & F1 (\%) ($\uparrow$) \\
\midrule
CMC      & 89.53 \mygray{$\pm$ 7.62} & 89.33 \mygray{$\pm$ 7.78} & 94.25 \mygray{$\pm$ 6.67}  & 10.38 \mygray{$\pm$ 0.63} & 74.97 \mygray{$\pm$ 1.23} & 74.82 \mygray{$\pm$ 2.18} \\
Cosmo    & 94.21 \mygray{$\pm$ 0.50} & 94.04 \mygray{$\pm$ 0.54} & 98.24 \mygray{$\pm$ 13.77} & 10.44 \mygray{$\pm$ 0.83} & 84.37 \mygray{$\pm$ 0.33} & 85.30 \mygray{$\pm$ 0.43} \\
SimCLR   & 95.53 \mygray{$\pm$ 0.73} & 95.41 \mygray{$\pm$ 0.74} & 99.87 \mygray{$\pm$ 11.31} & 10.29 \mygray{$\pm$ 0.52} & 84.36 \mygray{$\pm$ 0.47} & 85.49 \mygray{$\pm$ 0.36} \\
AudioMAE & 99.06 \mygray{$\pm$ 0.23} & 99.03 \mygray{$\pm$ 0.24} & 33.65 \mygray{$\pm$ 3.51}  &  5.65 \mygray{$\pm$ 0.29} & 89.18 \mygray{$\pm$ 0.32} & 90.11 \mygray{$\pm$ 0.53} \\
CAV-MAE  & 98.97 \mygray{$\pm$ 0.04} & 98.94 \mygray{$\pm$ 0.04} & 31.58 \mygray{$\pm$ 3.57}  &  5.48 \mygray{$\pm$ 0.37} & 88.12 \mygray{$\pm$ 0.24} & 89.05 \mygray{$\pm$ 0.35} \\
FOCAL    & 93.62 \mygray{$\pm$ 0.75} & 93.46 \mygray{$\pm$ 0.76} & 131.50 \mygray{$\pm$ 1.48} & 12.53 \mygray{$\pm$ 0.09} & 84.98 \mygray{$\pm$ 0.73} & 86.24 \mygray{$\pm$ 0.77} \\
FreqMAE  & 92.72 \mygray{$\pm$ 0.75} & 92.55 \mygray{$\pm$ 0.79} & 54.08 \mygray{$\pm$ 5.44}  &  7.14 \mygray{$\pm$ 0.25} & 83.43 \mygray{$\pm$ 0.56} & 84.07 \mygray{$\pm$ 0.50} \\
PhyMask  & 83.38 \mygray{$\pm$ 2.33} & 82.68 \mygray{$\pm$ 2.27} & 56.39 \mygray{$\pm$ 3.27}  &  7.67 \mygray{$\pm$ 0.39} & 84.79 \mygray{$\pm$ 3.23} & 82.15 \mygray{$\pm$ 9.13} \\
\rowcolor{lightblue}
\MODEL   & \textbf{99.27 \mygray{$\pm$ 0.07}} & \textbf{99.26 \mygray{$\pm$ 0.07}} 
         & \textbf{23.46 \mygray{$\pm$ 2.77}} & \textbf{5.37 \mygray{$\pm$ 0.24}} 
         & \textbf{89.63 \mygray{$\pm$ 0.57}} & \textbf{90.45 \mygray{$\pm$ 0.63}} \\
\bottomrule
\end{tabular}
\end{adjustbox}
\end{center}
\vspace{-0.1in}
\end{table*}

\subsubsection{Single-vehicle Localization.} 
Site H24\footnote{Appendix~2 summarizes site definitions from~\citep{li2025restoreml} explaining characteristics such as locations, terrain, and weather.}  was used for task-specific fine-tuning; a single-layer transformer head was tuned. The task is to predict each vehicle's ground-plane position from synchronized acoustic and vibration signals, with labeled supervision varied from $100\%$ to $20\%$ of site data. Table~\ref{tb:m3n-vc1} reports the mean square error (MSE) and average distance error: \MODEL{} is best at every label ratio. At full supervision, it cuts MSE from $18.85$ (CAV-MAE, the strongest baseline) to $12.98$---a $31\%$ improvement---and at $20\%$ supervision still attains $21.36$\,$m^2$, confirming that reconstruction objectives transfer most of the inductive bias into the encoder, so that the downstream task head can be trained effectively under scarce supervision.

\subsubsection{Single-vehicle Classification.} Here, we classify each sample into one of four vehicle types or a fifth ``background'' class, with a simple linear-classifier on site H24 under the cross-entropy loss. Table~\ref{tb:all} (M3N-VC block) reports accuracy and F1. \MODEL{} attains $99.27\%$ accuracy and $99.26\%$ F1, best among all methods. The confusion matrix in Figure~\ref{fig:cls_vehicle} (Appendix~4) shows that \MODEL{}'s residual errors are concentrated at the boundary of physically similar classes. This well-behaved error structure further supports the performance gains on the downstream task. Similar results were obtained for other sites.

\subsubsection{Multi-vehicle Joint Classification and Localization.} 
\begin{figure}[!tb]
    \centering
    \includegraphics[width=\linewidth]{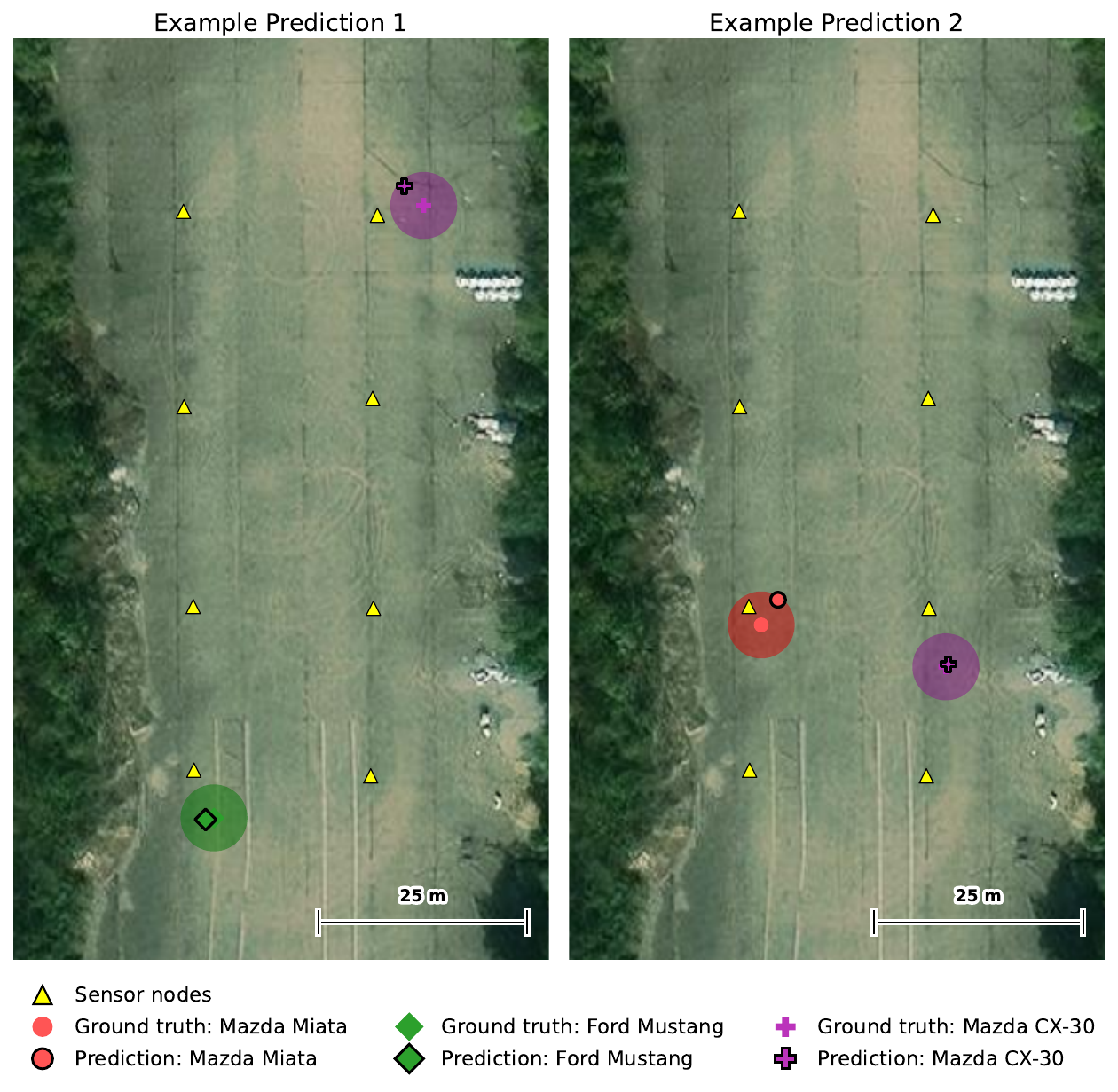}
    \vspace{-0.2in}
    \caption{\small Two representative SPAR predictions on the multi-vehicle joint recognition and localization task (site I22). Each panel displays sensor positions (yellow triangles), ground-truth vehicle classes and locations (filled markers, color-coded by class), and SPAR's predicted classes and locations (hollow markers of matching shape). The 4-meter radius drawn around each ground-truth vehicle marks the spatial threshold used by the mAP@4m metric. In both panels, every prediction matches the ground-truth class and falls well inside the 4-meter radius. Basemap imagery: \copyright~Esri, Vantor, Earthstar Geographics, and the GIS User Community.}
    \label{fig:multi}
\vspace{-0.15in}
\end{figure}

In this task, the goal is to enumerate every vehicle present in a scene and return its type and position, given only acoustic and seismic data. In (say) YOLO~\cite{terven2023comprehensive}, such a task would return bounding boxes around respective targets. Our acoustic-and-seismic-signal-based task is harder than its visual counterpart, because our modalities propagate \emph{omnidirectionally}, so signals from co-located sources superpose at every sensor, and the model must learn to recover each source's type and position from the resulting mixture rather than from spatially distinct regions of the input. We fine-tune the task-specific head on site I22 data, the only site that contains multiple simultaneously moving vehicles. The head comprises a two-layer transformer trained under a DETR-style set-prediction loss~\citep{carion2020end} against noisy 1\,Hz smartphone GPS labels. Table~\ref{tb:m3n-vc2} reports mAP@\emph{r} for $r \in \{4, 6, 8, 10\}$\,m, counting a prediction correct only when both the type and the position (within radius $r$) are correct. \MODEL{} reaches significantly higher mAP at all distance thresholds. Nevertheless, mAP@\emph{r} falls off sharply once a method's typical localization error exceeds $r$, therefore the large gap between \MODEL{} and baselines is a sign that the baselines' spatial precision is limited at tight thresholds, rather than a sign that the baselines collapse on this task. Figure~\ref{fig:multi} shows two representative SPAR predictions; predicted classes and positions closely match the ground truth in both cases. The observations confirm that our placement-aware representation improves the quality of downstream spatial analytics without degrading other tasks such as target classification. Importantly, it supports joint reasoning about target types and locations, a challenge that is generally hard to address with omnidirectional sensors. 

\subsection{Independence of Sensor Placement} 
The requirement that encoder training be independent from future sensor placement entails that sensors may be placed in configurations not seen at training time. To test whether the pretrained encoder transfers to such a placement, in this section, sites H08 and H24 (which share similar node configurations) are held out from encoder pretraining; the task head is then fine-tuned and evaluated on site H24 (now not previously seen in pre-training data). To avoid any further pretraining on a held-out site, each of its sensors is simply assigned the calibration token of a randomly chosen sensor seen in pretraining. Table~\ref{tb:m3n-vc5} shows the results. The table shows that \MODEL{} degrades gracefully relative to its in-distribution result in Table~\ref{tb:m3n-vc1} and still leads every baseline, though by a narrower margin, confirming robustness to unseen placement. To assess this on a second layout and on the harder task, we further hold out site I22 from pretraining and fine-tune and evaluate the multi-vehicle joint recognition and localization head on it. Table~\ref{tb:m3n-vc5b} shows that \MODEL{} again degrades relative to its in-distribution result in Table~\ref{tb:m3n-vc2} yet remains far ahead of every baseline, confirming that the encoder transfers to unseen placements across sites and tasks. Note that the randomly assigned tokens do not match the held-out sensors' actual characteristics, which may account for part of the degradation. \MODEL{} leads the baselines even under this simple strategy.

\begin{table}[!tb]
\caption{\small Generalization to a second placement not observed during pretraining (M3N-VC site I22 held out from pretraining; multi-vehicle head fine-tuned and evaluated on I22); mAP@\emph{r} (\%) ($\uparrow$).}
\vspace{-0.1in}
\label{tb:m3n-vc5b}
\begin{center}
\begin{adjustbox}{max width=\linewidth}
\begin{tabular}{lcccc}
\toprule
\multirow{2}{*}{Method} & \multicolumn{4}{c}{Held-out I22: Multi-vehicle Joint Recognition and Localization} \\ \cmidrule(lr){2-5}
                        & \multicolumn{1}{c}{mAP@4m} & \multicolumn{1}{c}{mAP@6m} & \multicolumn{1}{c}{mAP@8m} & \multicolumn{1}{c}{mAP@10m} \\
\midrule
CMC                     & \multicolumn{1}{c}{0.02 \mygray{$\pm$ 0.03}} & \multicolumn{1}{c}{0.19 \mygray{$\pm$ 0.24}} & \multicolumn{1}{c}{0.81 \mygray{$\pm$ 0.83}} & \multicolumn{1}{c}{2.04 \mygray{$\pm$ 1.74}} \\
Cosmo                   & \multicolumn{1}{c}{0.26 \mygray{$\pm$ 0.14}} & \multicolumn{1}{c}{1.54 \mygray{$\pm$ 0.61}} & \multicolumn{1}{c}{4.43 \mygray{$\pm$ 1.43}} & \multicolumn{1}{c}{8.77 \mygray{$\pm$ 2.92}} \\
SimCLR                  & \multicolumn{1}{c}{0.32 \mygray{$\pm$ 0.05}} & \multicolumn{1}{c}{1.97 \mygray{$\pm$ 0.32}} & \multicolumn{1}{c}{5.55 \mygray{$\pm$ 0.51}} & \multicolumn{1}{c}{10.56 \mygray{$\pm$ 1.14}} \\
AudioMAE                & \multicolumn{1}{c}{0.99 \mygray{$\pm$ 0.39}} & \multicolumn{1}{c}{5.06 \mygray{$\pm$ 1.98}} & \multicolumn{1}{c}{11.98 \mygray{$\pm$ 3.27}} & \multicolumn{1}{c}{20.85 \mygray{$\pm$ 3.89}} \\
CAV-MAE                 & \multicolumn{1}{c}{2.70 \mygray{$\pm$ 0.30}} & \multicolumn{1}{c}{10.83 \mygray{$\pm$ 0.77}} & \multicolumn{1}{c}{24.06 \mygray{$\pm$ 2.08}} & \multicolumn{1}{c}{37.71 \mygray{$\pm$ 2.90}} \\
FOCAL                   & \multicolumn{1}{c}{0.18 \mygray{$\pm$ 0.10}} & \multicolumn{1}{c}{1.38 \mygray{$\pm$ 0.30}} & \multicolumn{1}{c}{4.23 \mygray{$\pm$ 0.67}} & \multicolumn{1}{c}{8.88 \mygray{$\pm$ 0.89}} \\
FreqMAE                 & \multicolumn{1}{c}{0.64 \mygray{$\pm$ 0.16}} & \multicolumn{1}{c}{3.81 \mygray{$\pm$ 0.43}} & \multicolumn{1}{c}{9.69 \mygray{$\pm$ 1.22}} & \multicolumn{1}{c}{16.92 \mygray{$\pm$ 2.19}} \\
PhyMask                 & \multicolumn{1}{c}{0.30 \mygray{$\pm$ 0.04}} & \multicolumn{1}{c}{2.22 \mygray{$\pm$ 0.14}} & \multicolumn{1}{c}{5.91 \mygray{$\pm$ 0.11}} & \multicolumn{1}{c}{11.42 \mygray{$\pm$ 0.69}} \\ \rowcolor{lightblue}
\MODEL                  & \multicolumn{1}{c}{\textbf{19.43 \mygray{$\pm$ 3.77}}} & \multicolumn{1}{c}{\textbf{49.46 \mygray{$\pm$ 6.46}}} & \multicolumn{1}{c}{\textbf{70.37 \mygray{$\pm$ 6.33}}} & \multicolumn{1}{c}{\textbf{82.10 \mygray{$\pm$ 3.51}}} \\
\bottomrule
\end{tabular}
\end{adjustbox}
\end{center}
\vspace{-0.1in}
\end{table}

\subsection{Component Necessity: An Ablation Study}\label{exp:ablation} 
Since the main contribution of our approach is to improve {\em spatial\/} representation, we center the ablation study specifically on a task that focuses on spatial analytics. Namely, we explore the contributions of different design choices to downstream {\em localization accuracy\/}.
In the following subsections, we first confirm the key contribution of \MODEL{} -- that pre-training an encoder to create a {\em latent representation that jointly embeds positional information and signals\/} is better than simply associating location metadata with signal embeddings. We then ablate over other design choices, most importantly the presence of {\em per-sensor calibration tokens\/} and the details of the training loss function.  

\begin{figure}[!tb]
\centering
\includegraphics[width=0.75\linewidth]{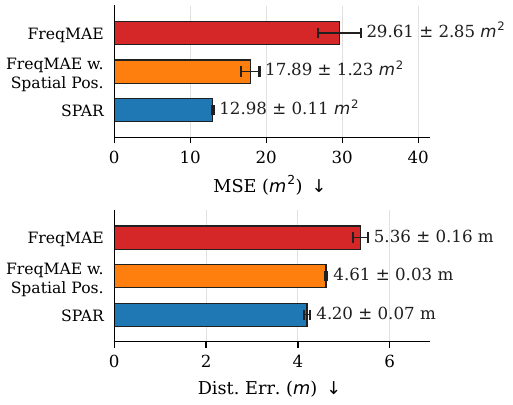}
\vspace{-0.18in}
\caption{\small Isolating the role of pretraining on M3N-VC single-vehicle localization at the full label ratio. Augmenting FreqMAE with the same spatial positional information \MODEL{} uses---added at fine-tuning via a learnable projection---closes part of the gap to \MODEL{}, showing that \MODEL{}'s spatial-symmetric pretraining strategy is itself a meaningful source of its advantage.}
\Description{Two vertically stacked panels of horizontal bars. The top panel shows MSE in meters squared; the bottom panel shows distance error in meters. Each panel has three colored bars (FreqMAE in red, +Spatial in orange, and SPAR in blue) sticking together edge-to-edge with method names on the left y-axis. Mean and standard deviation are annotated to the right of each bar with their unit.}
\label{fig:freqmae_spatial}
\vspace{-0.2in}
\end{figure}

\subsubsection{Location Embedding versus Location Metadata Access.} 
A natural concern is whether \MODEL{}'s advantage over placement-agnostic baselines stems simply from \emph{access} to sensor positions rather than from the way these (deployment-time) positions are translated into the latent space by the pretrained encoder. We construct a strengthened variant of FreqMAE~\citep{kara2024freqmae}, \MODEL{}'s closest architectural cousin among the baselines, and grant it the same spatial positional information, but only at fine-tuning time (since its pre-training is location-agnostic).\footnote{We fine-tune both algorithms at full supervision in order to eliminate artifacts arising from the choice of data.} We employ a learnable linear layer to project each sensor's position into the latent-embedding dimension and add the resulting embeddings to FreqMAE's latent embeddings before the localization head, leaving FreqMAE's pretraining objective unchanged. Figure~\ref{fig:freqmae_spatial} shows that fine-tune-only access to sensor positions is genuinely useful, but \MODEL{} is able to enhance the performance further on top of it, confirming that \MODEL{}'s performance gain is attributed not merely to access to location information but also to learning to represent this information effectively in the latent space thanks to the proposed novel encoder pretraining.

\begin{figure}[!tb]
\centering
\includegraphics[width=0.85\linewidth]{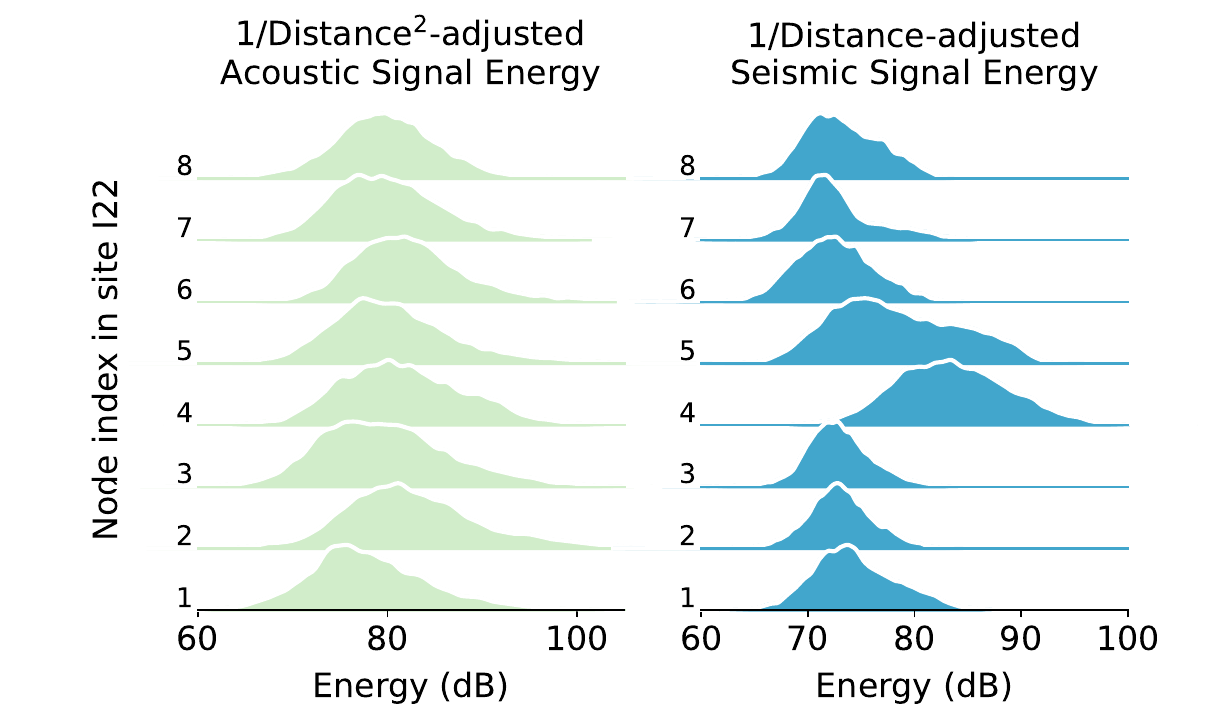}
\vspace{-0.15in}
\caption{\small Distance-corrected signal-energy distributions of the eight nodes at M3N-VC site I22, acoustic (left) and seismic (right).}
\Description{Two panels of ridge plots of signal energy in decibels, one ridge per node for the eight nodes of site I22, acoustic on the left and seismic on the right; the ridges differ visibly in center and width.}
\label{fig:HL_ridge}
\vspace{-0.15in}
\end{figure}

\subsubsection{Design Ablations.}
Beyond the above key experiment, Table~\ref{tb:m3n-vc6} quantifies the contributions of different components of our design to localization accuracy. Removing the per-sensor calibration embedding produces the largest degradation ($+74\%$ MSE), confirming that the encoder relies on calibration as conditioning context. Removing the spatial reconstruction objective, the spatial positional embedding, or the geometric augmentation each produces a smaller but jointly comparable drop ($+13$--$20\%$ MSE), indicating that these three components are complementary rather than redundant. For the masking strategy, Sensor-Drop Masking (masking entire sensors) degrades MSE noticeably, while Sensor-Balanced Masking (a minimum number of unmasked tokens per sensor) slightly improves over the default. The framework is otherwise insensitive to the mask ratio: varying it from $0.5$ to $0.85$ yields marginal changes to the task performance.

The size of the calibration effect reflects real heterogeneity: Figure~\ref{fig:HL_ridge} shows that even nominally identical nodes at one site differ visibly in signal-energy distribution. Under an MSE objective, such persistent per-sensor differences produce large residuals at unchanged content, which the token plausibly lets the encoder absorb rather than carry in the representation.

\begin{figure}[!tb]
\centering
\includegraphics[width=0.7\linewidth]{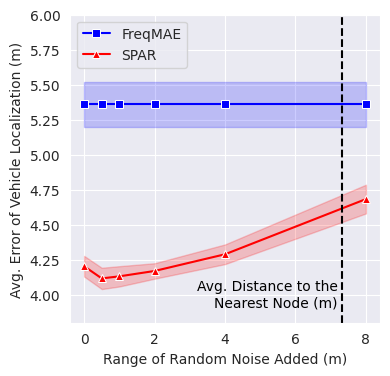}
\vspace{-0.15in}
\caption{\small Single-vehicle localization error on M3N-VC versus the range of uniform noise added to the position of one random node.}
\Description{Line plot of average vehicle localization error in meters against the range of random position noise in meters, from 0 to 8. The SPAR curve rises gradually from about 4.2 to 4.7 meters; the FreqMAE curve is flat at about 5.4 meters. A dashed vertical line marks the average distance to the nearest node near 7.4 meters.}
\label{fig:noise_vehicle_localization}
\vspace{-0.15in}
\end{figure}

\begin{table}[!htb]
\caption{\small Single-vehicle localization on M3N-VC under per-node message drop (drop rate $0.1$ and $0.2$).}
\vspace{-0.1in}
\label{tb:m3n-vc4}
\begin{center}
\begin{adjustbox}{max width=\linewidth}
\begin{tabular}{lcccc}
\toprule
\multirow{3}{*}{Method} & \multicolumn{4}{c}{M3N-VC Single-vehicle Localization} \\ \cmidrule(lr){2-5}
                        & \multicolumn{2}{c}{Message Drop Rate 0.1} & \multicolumn{2}{c}{Message Drop Rate 0.2} \\ \cmidrule(lr){2-3} \cmidrule(lr){4-5}
                        & \multicolumn{1}{c}{MSE ($m^2$) ($\downarrow$)} & \multicolumn{1}{c}{Dist. Err. ($m$) ($\downarrow$)} & \multicolumn{1}{c}{MSE ($m^2$) ($\downarrow$)} & \multicolumn{1}{c}{Dist. Err. ($m$) ($\downarrow$)} \\
\midrule
CMC                     & \multicolumn{1}{c}{50.27 \mygray{$\pm$ 18.03}} & \multicolumn{1}{c}{6.60 \mygray{$\pm$ 1.17}} & \multicolumn{1}{c}{55.38 \mygray{$\pm$ 22.01}} & \multicolumn{1}{c}{6.91 \mygray{$\pm$ 1.29}} \\
Cosmo                   & \multicolumn{1}{c}{29.37 \mygray{$\pm$ 2.43}} & \multicolumn{1}{c}{5.44 \mygray{$\pm$ 0.10}} & \multicolumn{1}{c}{33.19 \mygray{$\pm$ 2.67}} & \multicolumn{1}{c}{5.61 \mygray{$\pm$ 0.12}} \\
SimCLR                  & \multicolumn{1}{c}{25.65 \mygray{$\pm$ 1.28}} & \multicolumn{1}{c}{5.15 \mygray{$\pm$ 0.07}} & \multicolumn{1}{c}{28.31 \mygray{$\pm$ 3.94}} & \multicolumn{1}{c}{5.4 \mygray{$\pm$ 0.18}} \\
AudioMAE                & \multicolumn{1}{c}{18.25 \mygray{$\pm$ 1.23}} & \multicolumn{1}{c}{4.77 \mygray{$\pm$ 0.10}} & \multicolumn{1}{c}{19.03 \mygray{$\pm$ 1.14}} & \multicolumn{1}{c}{4.77 \mygray{$\pm$ 0.05}} \\
CAV-MAE                 & \multicolumn{1}{c}{15.85 \mygray{$\pm$ 0.81}} & \multicolumn{1}{c}{4.57 \mygray{$\pm$ 0.10}} & \multicolumn{1}{c}{15.98 \mygray{$\pm$ 1.67}} & \multicolumn{1}{c}{4.44 \mygray{$\pm$ 0.12}} \\
FOCAL                   & \multicolumn{1}{c}{27.42 \mygray{$\pm$ 2.33}} & \multicolumn{1}{c}{5.32 \mygray{$\pm$ 0.20}} & \multicolumn{1}{c}{33.03 \mygray{$\pm$ 2.28}} & \multicolumn{1}{c}{5.70 \mygray{$\pm$ 0.15}} \\
FreqMAE                 & \multicolumn{1}{c}{28.32 \mygray{$\pm$ 1.22}} & \multicolumn{1}{c}{5.17 \mygray{$\pm$ 0.14}} & \multicolumn{1}{c}{28.68 \mygray{$\pm$ 6.96}} & \multicolumn{1}{c}{5.32 \mygray{$\pm$ 0.24}} \\
PhyMask                 & \multicolumn{1}{c}{24.09 \mygray{$\pm$ 3.31}} & \multicolumn{1}{c}{5.03 \mygray{$\pm$ 0.27}} & \multicolumn{1}{c}{27.37 \mygray{$\pm$ 2.04}} & \multicolumn{1}{c}{5.31 \mygray{$\pm$ 0.21}} \\ \rowcolor{lightblue}
\MODEL                  & \multicolumn{1}{c}{\textbf{12.48 \mygray{$\pm$ 0.68}}} & \multicolumn{1}{c}{\textbf{4.07 \mygray{$\pm$ 0.02}}} & \multicolumn{1}{c}{\textbf{14.56 \mygray{$\pm$ 2.47}}} & \multicolumn{1}{c}{\textbf{4.27 \mygray{$\pm$ 0.07}}} \\
\bottomrule
\end{tabular}
\end{adjustbox}
\end{center}
\vspace{-0.15in}
\end{table}

\subsubsection{Robustness to Imperfect Spatial Information.}
We next perturb the spatial input itself. First, the position of one randomly chosen node is corrupted with uniform random noise of increasing range during both fine-tuning and testing. Figure~\ref{fig:noise_vehicle_localization} shows that the localization error of \MODEL{} grows smoothly with the noise range and remains below that of the placement-agnostic FreqMAE even when the range exceeds the average distance to the nearest node. Second, each node's data is independently dropped with probability $0.1$ or $0.2$ during fine-tuning and evaluation. Table~\ref{tb:m3n-vc4} shows that \MODEL{} retains the lowest MSE and distance error at every drop rate. The representation thus depends on spatial information, as intended, while degrading gracefully when that information is noisy or incomplete.

\begin{table}[!tb]
\caption{\small Cosine similarity of \MODEL{} representations for 128 sample pairs of each kind (M3N-VC sites H08 and H24).}
\vspace{-0.1in}
\label{tb:consist}
\begin{center}
\begin{adjustbox}{max width=\linewidth}
\begin{tabular}{lccc}
\toprule
Pairs & \multicolumn{1}{c}{Matched H08--H24} & \multicolumn{1}{c}{Random within H08} & \multicolumn{1}{c}{Random within H24} \\
\midrule
Cosine similarity & \multicolumn{1}{c}{0.934 \mygray{$\pm$ 0.038}} & \multicolumn{1}{c}{0.815 \mygray{$\pm$ 0.052}} & \multicolumn{1}{c}{0.806 \mygray{$\pm$ 0.063}} \\
\bottomrule
\end{tabular}
\end{adjustbox}
\end{center}
\vspace{-0.1in}
\end{table}

\subsubsection{Consistency across Sensor Conditions.}
We then test whether the representation is robust to changing sensor conditions. Sites H08 and H24 share a node layout but differ in weather and hence in per-sensor conditions. We form 128 \emph{matched} pairs, each consisting of one sample from H08 and one from H24 with the same vehicle class and a vehicle location within 1\,m, and compare the cosine similarity of their representations against that of 128 \emph{random} pairs drawn within H08 and 128 drawn within H24. Table~\ref{tb:consist} shows that matched pairs across sites are markedly closer than random pairs within a site, indicating that the representation follows scene content and is robust to the change in sensor conditions.

\subsection{Run-time Overhead and Field Testing}\label{exp:casestudy}

We then test \MODEL{} in an outdoors setting by reconstituting the vehicle-monitoring task of Section~\ref{exp:m3n-vc} on a new outdoor site. Beyond task performance, we measure the two operational constraints any fielded distributed-sensing system must clear: per-sample inference cost on commodity edge hardware, and per-sample network footprint on a bandwidth-constrained network.

\begin{figure}[!tb]
\centering
\includegraphics[width=\linewidth]{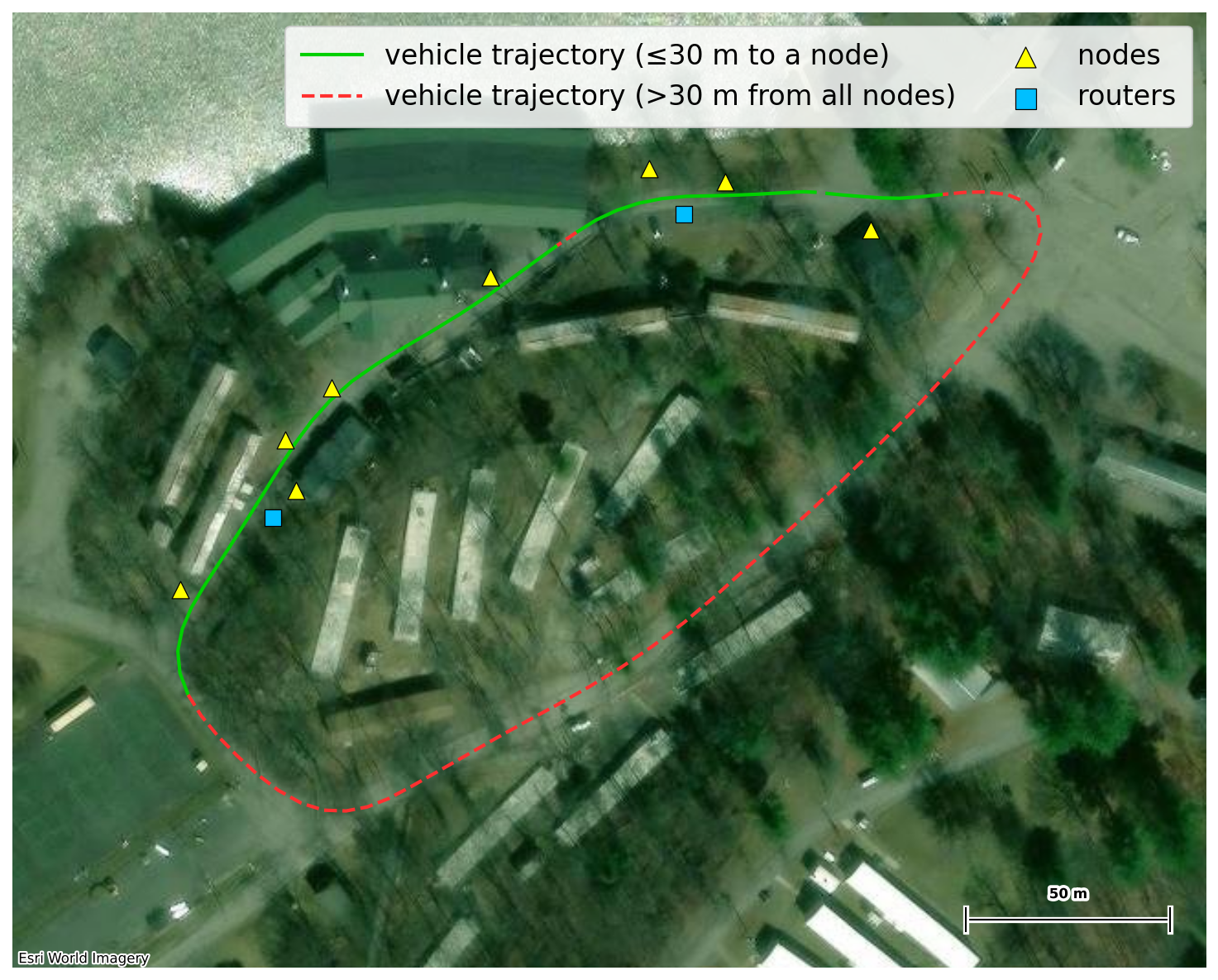}
\vspace{-0.25in}
\caption{\small Case-study site. Eight sensor nodes (yellow triangles) are deployed across an outdoor field. The vehicle trajectory is split into a foreground segment (green solid) within 30\,m of at least one node and a background segment (red dashed) beyond 30\,m of every node, following the labeling convention adopted in this study. Basemap imagery: \copyright~Esri, Vantor, Earthstar Geographics, and the GIS User Community.}
\Description{Satellite view of the case-study site. Yellow triangles mark eight sensor nodes. A green solid line marks the portion of the vehicle trajectory within 30 meters of at least one node (foreground, labeled with vehicle type and location), and a red dashed line marks the portion beyond 30 meters from all nodes (labeled as background). A 30-meter scale bar is shown in the lower-right.}
\label{fig:wp_nodes}
\vspace{-0.15in}
\end{figure}

\begin{figure}[!tb]
\centering
\includegraphics[width=\linewidth]{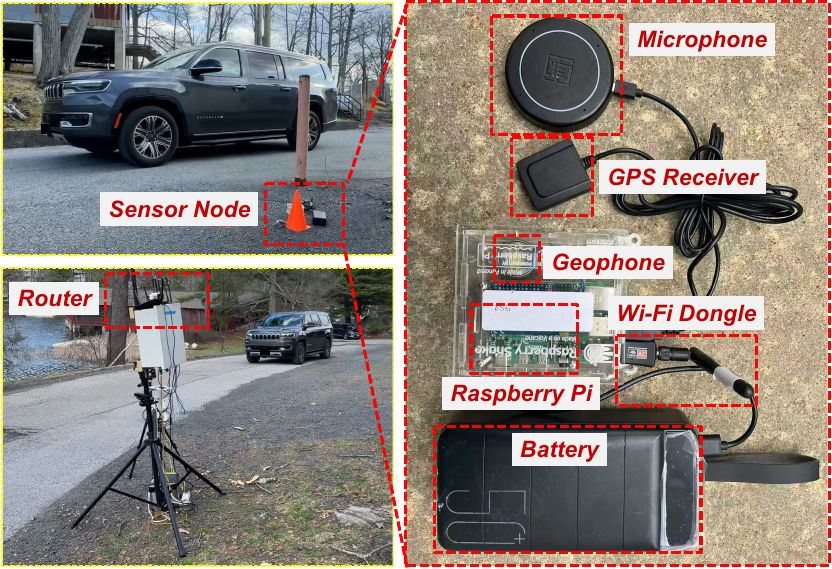}
\vspace{-0.25in}
\caption{\small Field hardware for the case study. \emph{Left}: a deployed sensor node in situ (top) and the tripod-mounted Wi-Fi router that backhauls the array (bottom). \emph{Right}: the internal components of each node---a microphone, a geophone, a GPS receiver, a Raspberry~Pi controller, a Wi-Fi dongle, and a battery. The eight nodes are identical custom builds installed for the full duration of the study.}
\Description{Composite photograph of the field-deployed sensor hardware. Top-left: a sensor node mounted on a tripod beside a vehicle. Bottom-left: a Wi-Fi router on a tripod in the outdoor field. Right: a close-up annotated view of an opened sensor node, with six labeled components: microphone, geophone, GPS receiver, Raspberry Pi, Wi-Fi dongle, and battery.}
\label{fig:wp_setup}
\vspace{-0.15in}
\end{figure}

\begin{table}[!tb]
\caption{\small Case-study specifications.}
\vspace{-0.1in}
\label{tb:casestudy_specs}
\begin{center}
\begin{adjustbox}{max width=0.8\linewidth}
\begin{tabular}{lc}
\toprule
                       & Case Study \\
\midrule
Spatial scale          & $100$\,m-radius \\
Modalities             & microphone $+$ geophone \\
Sampling rate          & $16$\,kHz / $200$\,Hz \\
Nodes per sample       & $8$ \\
Sample length          & $2$\,s \\
Total samples          & $6{,}427$ \\
Train / val split      & temporal ($\approx$ $3{:}1$) \\
Downstream task        & vehicle joint recognition and localization \\
\bottomrule
\end{tabular}
\end{adjustbox}
\end{center}
\vspace{-0.15in}
\end{table}

\noindent\textbf{Setup.}
The site is an outdoor field instrumented with eight identical, self-contained multi-sensor nodes. Each node integrates a microphone, a geophone, a GPS receiver (for time-synchronization across the array), a Raspberry~Pi controller, a Wi-Fi dongle, and a battery into a single enclosure, with a tripod-mounted Wi-Fi router backhauling the array (Figure~\ref{fig:wp_setup}). Table~\ref{tb:casestudy_specs} summarizes the specifications, which resemble the M3N-VC configurations except for the site, the specific geometric layout, and per-sensor calibration factors, and the vehicle types. The sensor placements and representative vehicle trajectories are overlaid on the satellite view in Figure~\ref{fig:wp_nodes}. Two vehicles from distinct categories---a full-size Jeep Wagoneer and a subcompact Hyundai Kona---are driven over the site one at a time, in both directions and at varying speeds across the indicated route. The encoder is initialized from the M3N-VC pretrained checkpoint and further pretrained, without labels, on the on-site data, so the calibration tokens of the field nodes are learned during pretraining; the study is thus not intended to test generalization to unseen placements.
Table~\ref{tb:casestudy_map} reports mAP@\emph{r} against all baselines. \MODEL{} leads at every threshold. We now turn to inference and bandwidth {\em cost\/}.

\begin{table}[!tb]
\caption{\small Vehicle joint recognition and localization on the case-study site; mAP@\emph{r} (\%) ($\uparrow$) with spatial tolerance $r \in \{8,12,16,20\}$~m.}
\vspace{-0.1in}
\label{tb:casestudy_map}
\begin{center}
\begin{adjustbox}{max width=\linewidth}
\begin{tabular}{lcccc}
\toprule
\multirow{2}{*}{Method} & \multicolumn{4}{c}{Case Study: Joint Recognition and Localization} \\ \cmidrule(lr){2-5}
                        & \multicolumn{1}{c}{mAP@8m} & \multicolumn{1}{c}{mAP@12m} & \multicolumn{1}{c}{mAP@16m} & \multicolumn{1}{c}{mAP@20m} \\
\midrule
CMC                     & \multicolumn{1}{c}{7.53 \mygray{$\pm$ 0.76}} & \multicolumn{1}{c}{16.95 \mygray{$\pm$ 2.91}} & \multicolumn{1}{c}{27.68 \mygray{$\pm$ 3.44}} & \multicolumn{1}{c}{36.24 \mygray{$\pm$ 3.34}} \\
Cosmo                   & \multicolumn{1}{c}{12.05 \mygray{$\pm$ 0.98}} & \multicolumn{1}{c}{25.57 \mygray{$\pm$ 1.95}} & \multicolumn{1}{c}{36.85 \mygray{$\pm$ 1.93}} & \multicolumn{1}{c}{47.12 \mygray{$\pm$ 3.12}} \\
SimCLR                  & \multicolumn{1}{c}{12.23 \mygray{$\pm$ 1.50}} & \multicolumn{1}{c}{25.40 \mygray{$\pm$ 3.35}} & \multicolumn{1}{c}{38.09 \mygray{$\pm$ 4.05}} & \multicolumn{1}{c}{48.89 \mygray{$\pm$ 4.03}} \\
AudioMAE                & \multicolumn{1}{c}{15.41 \mygray{$\pm$ 0.97}} & \multicolumn{1}{c}{32.22 \mygray{$\pm$ 0.82}} & \multicolumn{1}{c}{48.09 \mygray{$\pm$ 1.83}} & \multicolumn{1}{c}{60.48 \mygray{$\pm$ 2.69}} \\
CAV-MAE                 & \multicolumn{1}{c}{34.24 \mygray{$\pm$ 3.19}} & \multicolumn{1}{c}{58.80 \mygray{$\pm$ 5.55}} & \multicolumn{1}{c}{73.76 \mygray{$\pm$ 3.22}} & \multicolumn{1}{c}{80.42 \mygray{$\pm$ 2.58}} \\
FOCAL                   & \multicolumn{1}{c}{12.03 \mygray{$\pm$ 0.87}} & \multicolumn{1}{c}{24.95 \mygray{$\pm$ 2.54}} & \multicolumn{1}{c}{37.47 \mygray{$\pm$ 1.75}} & \multicolumn{1}{c}{48.43 \mygray{$\pm$ 3.00}} \\
FreqMAE                 & \multicolumn{1}{c}{9.13 \mygray{$\pm$ 2.07}} & \multicolumn{1}{c}{19.64 \mygray{$\pm$ 3.72}} & \multicolumn{1}{c}{33.45 \mygray{$\pm$ 5.91}} & \multicolumn{1}{c}{44.66 \mygray{$\pm$ 7.78}} \\
PhyMask                 & \multicolumn{1}{c}{9.70 \mygray{$\pm$ 0.60}} & \multicolumn{1}{c}{22.44 \mygray{$\pm$ 2.90}} & \multicolumn{1}{c}{34.59 \mygray{$\pm$ 2.77}} & \multicolumn{1}{c}{44.63 \mygray{$\pm$ 2.44}} \\ \rowcolor{lightblue}
\MODEL                  & \multicolumn{1}{c}{\textbf{41.57 \mygray{$\pm$ 5.42}}} & \multicolumn{1}{c}{\textbf{68.33 \mygray{$\pm$ 4.41}}} & \multicolumn{1}{c}{\textbf{80.22 \mygray{$\pm$ 1.22}}} & \multicolumn{1}{c}{\textbf{86.19 \mygray{$\pm$ 0.49}}} \\
\bottomrule
\end{tabular}
\end{adjustbox}
\end{center}
\vspace{-0.1in}
\end{table}

\subsubsection{Inference cost.}
\MODEL{} comprises 205.3M parameters and executes 2.78~GFLOPs per fused multi-modal sample. The per-sensor calibration tokens account for $Knd_\mR$ of these parameters for $n$ nodes with $K$ modalities ($4Knd_\mR$ bytes at fp32)---with $K=2$ and $d_\mR=24$, 19.2~KB for 100 nodes and 192~KB for 1,000---while the projection layer consuming them is fixed in size. Unbatched per-sample latency is 7.83~ms on an NVIDIA RTX~3090 (representative of a co-located server) and 27.64~ms on an Apple M2~Pro Mac~mini with 32~GB of RAM (representative of a fanless edge gateway), which can support over 35 inferences per second on the Mac~mini alone, well above the rate at which the underlying signals are produced. \MODEL{}'s task performance is therefore deliverable in real time on the kind of unobtrusive hardware a field operator could reasonably install.

\begin{figure}[!tb]
\centering
\includegraphics[width=0.85\linewidth]{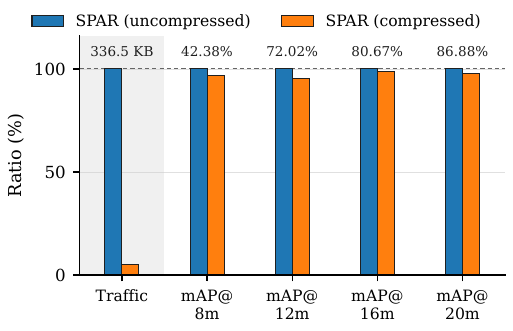}
\vspace{-0.18in}
\caption{\small Impact of JPEG compression on \MODEL{} in the field deployment case study. Each of the five scenarios (per-sample traffic, and mAP at four spatial thresholds) is shown as a pre-compression/post-compression pair: the pre-compression bar is normalized to $100\%$, labeled above with its absolute value and unit; the post-compression bar is shown at its ratio to the pre-compression bar. The per-sample traffic shrinks from $336.48$\,KB to $17.54$\,KB ($5.21\%$ of the uncompressed payload, a $19.2\times$ reduction), while mAP drops by only $1$--$3$ percentage points across the four thresholds. This experiment uses a single seed.}
\Description{A single vertical-bar plot containing five pre/post pairs sticking together side by side: Traffic, mAP at 8 meters, mAP at 12 meters, mAP at 16 meters, and mAP at 20 meters. The pre bar is always at one hundred percent; the post bar sits in a light-gray vertical band and shows the ratio to the pre bar. The traffic post bar is dramatically shorter than the pre bar; the four mAP post bars are nearly indistinguishable from the corresponding pre bars.}
\label{tb:casestudy_compression}
\vspace{-0.15in}
\end{figure}

\subsubsection{Bandwidth cost.}
The remaining constraint is the node-to-aggregator link. Under an aggressive JPEG setting, the per-sample payload from each node shrinks from $336.48$\,KB to $17.54$\,KB on average---$5.21\%$ of the uncompressed size, a $19.2\times$ reduction. Figure~\ref{tb:casestudy_compression} reports mAP@\emph{r} on the compressed stream alongside the uncompressed baseline. The encoder is not retrained on compressed inputs, yet mAP falls by only $1$--$3$ percentage points across the four spatial thresholds. \MODEL{}'s representations therefore degrade gracefully under the standard lossy compression methods.
Taken together, \MODEL{} is able to run in real time on a Mac~mini-class edge device and can tolerate a near-twenty-fold network traffic reduction with only a $1$--$3$ percentage-point drop in mAP. We read this as concrete evidence of \MODEL{}'s deployment feasibility for distributed sensing applications in the field.

\subsection{Generalization to Other Applications}\label{sec:other_apps}

It remains to show that \MODEL{} is generalizable to other sensing applications besides vehicle tracking. Two other datasets are employed in this section for this purpose as discussed below. 

\subsubsection{Earthquake Monitoring.}
The Ridgecrest Seismicity Dataset~\citep{caltech1926socal} shifts the sensing regime in three operationally significant ways: the spatial scale jumps from meters to tens of kilometers, the station array is sparser, and weak seismic waves often fail to reach distant stations---leaving $20.38\%$ of inter-station channels empty per event. We build the dataset on the Southern California Seismic Network (SCSN)~\citep{caltech1926socal} catalog accessed via the Southern California Earthquake Data Center (SCEDC)~\citep{center2013southern}, following the methodology of Si~\etal~\citep{si2024all}: events from January~1, 2020 through December~31, 2024, with magnitude $> -0.5$ and depth $> -5$\,km, centered on $(35.9^\circ, -117.6^\circ)$. For each event, we collect three-component (East, North, Vertical) waveforms from 16 California Integrated Seismic Network stations in two co-located modalities---high-gain broadband seismometer and high-gain accelerometer---retrieved in miniSEED format from SCEDC Open Data. Each $30.72$-second window is detrended and converted into a spectrogram via the Short-Time Fourier Transform, with a $2$--$35$\,Hz band-pass filter suppressing oceanic and atmospheric microseisms below and instrumental and environmental noise above; station coordinates are converted to kilometer-level local positions via Local Tangent Plane projection~\citep{united1987department}. A year-based temporal split uses $2020$--$2021$ for training ($22{,}360$ events) and $2022$--$2024$ for validation ($9{,}092$ events).

The task is 3D earthquake localization from the 16-station multi-modal waveforms, using a two-layer transformer as the localization head and MSE as the loss function. The Ridgecrest block of Table~\ref{tb:all} reports MSE and average distance error: \MODEL{} attains $23.46$\,km$^2$ and $5.37$\,km, best on both metrics and a $26\%$ MSE improvement over the strongest baseline (CAV-MAE at $31.58$). For context, this accuracy, obtained with a frozen general-purpose encoder and a two-layer head, is of the same order as that reported by dedicated multi-station deep-learning locators from the geophysics literature: MSLOC~\citep{zhang2025multistation} attains a mean 3-D location error of about $5$\,km in Oklahoma and Southern California; TEAM-LM~\citep{munchmeyer2021earthquake} attains median epicentral errors of $2$--$14$\,km across Chile, Italy, and Japan; and PLAN~\citep{si2024all}, on the same Ridgecrest stations but under \emph{a random split} rather than a temporal split, attains a mean per-station offset error of $1.09$\,km and a mean depth error of $1.43$\,km. The effectiveness of \MODEL{} suggests that the design principles we adopted are more broadly applicable across distributed sensing applications. 

\begin{table}[!tb]
\caption{\small RealWorld-HAR activity recognition with one body sensor removed.}
\vspace{-0.1in}
\label{tb:har_drop}
\begin{center}
\begin{adjustbox}{max width=0.7\linewidth}
\begin{tabular}{lcc}
\toprule
Removed sensor & \multicolumn{1}{c}{Accuracy (\%) ($\uparrow$)} & \multicolumn{1}{c}{F1 (\%) ($\uparrow$)} \\
\midrule
None (Table~\ref{tb:all})  & \multicolumn{1}{c}{89.63 \mygray{$\pm$ 0.57}} & \multicolumn{1}{c}{90.45 \mygray{$\pm$ 0.63}} \\ \midrule
Head       & \multicolumn{1}{c}{89.03 \mygray{$\pm$ 0.21}} & \multicolumn{1}{c}{89.89 \mygray{$\pm$ 0.26}} \\
Chest      & \multicolumn{1}{c}{90.64 \mygray{$\pm$ 1.01}} & \multicolumn{1}{c}{91.38 \mygray{$\pm$ 1.04}} \\
Upper arm  & \multicolumn{1}{c}{89.66 \mygray{$\pm$ 0.93}} & \multicolumn{1}{c}{90.39 \mygray{$\pm$ 0.97}} \\
Waist      & \multicolumn{1}{c}{85.87 \mygray{$\pm$ 1.38}} & \multicolumn{1}{c}{87.00 \mygray{$\pm$ 1.42}} \\
Thigh      & \multicolumn{1}{c}{89.84 \mygray{$\pm$ 0.44}} & \multicolumn{1}{c}{90.79 \mygray{$\pm$ 0.53}} \\
Shin       & \multicolumn{1}{c}{85.79 \mygray{$\pm$ 1.08}} & \multicolumn{1}{c}{86.46 \mygray{$\pm$ 1.02}} \\
\bottomrule
\end{tabular}
\end{adjustbox}
\end{center}
\vspace{-0.2in}
\end{table}

\subsubsection{Human Activity Recognition.} A qualitatively different application is RealWorld-HAR~\citep{sztyler2016body}. It differs from the above two datasets in that (IMU) sensors are located on a single subject's body rather than spread across a field, so absolute spatial scale collapses to under a meter. The dataset records 15 participants performing eight activities (walking, sitting, lying, climbing down, climbing up, running, standing, jumping). Body-mounted nodes are originally instrumented at seven locations---chest, forearm, head, shin, thigh, upper arm, and waist---of which we retain six after dropping the forearm sensor due to substantial data loss. All sensor signals are resampled to $50$\,Hz and segmented into non-overlapping $4$-second windows; a subject-based split uses the first $10$ participants for training and the remaining $5$ for validation. Since the dataset provides no explicit spatial coordinates, we assign each sensor an approximate 3D position based on standard anatomical placement on a standing person. This ``spatial position'' of sensors, while coarse-grained, informs the encoder of the relative proximity of each sensor to the others, and proves empirically effective. The downstream head is a linear classifier trained under cross-entropy. The RealWorld-HAR block of Table~\ref{tb:all} reports accuracy and F1, showing that \MODEL{} remains the strongest method. The confusion matrix in Figure~\ref{fig:cls_activity} (Appendix~4) shows that \MODEL{}'s errors concentrate among activities with similar motion signatures. We further test tolerance to missing sensors: the head is fine-tuned once with one randomly chosen sensor removed per sample, then evaluated with each sensor in turn removed from every test sample. Table~\ref{tb:har_drop} shows that accuracy stays within one percentage point of the full-array result for four of the six sensors and drops by about four points when the waist or shin sensor is removed, confirming that a single fine-tuned head tolerates the loss of any one sensor. This section, while brief, offers empirical evidence that \MODEL{} is indeed generalizable across disparate sensing applications.

\section{Related Work}\label{sec:rw}
\noindent\textbf{Pretraining and Foundation Models for Sensing.}
Pretraining for sensing seeks transferable representations from unlabeled data and broadly follows three paradigms: contrastive learning, masked reconstruction, and LLM-based frameworks. Contrastive methods align multi-modal embeddings, ranging from intra-sample approaches such as Cosmo~\citep{ouyang2022cosmo}, Cocoa~\citep{deldari2022cocoa}, and FOCAL~\citep{liu2023focal}, to large-scale or loosely paired models like ImageBind~\citep{girdhar2023imagebind} and MMBind~\citep{ouyang2024mmbind}. Generative approaches are dominated by masked autoencoding~\citep{woo2024unified, das2024decoder}, including time-series adaptations such as Ti-MAE~\citep{li2023ti}, MOMENT~\citep{goswami2024moment}, TS-MAE~\citep{liu2025ts}, and frequency-aware variants like FreqMAE~\citep{kara2024freqmae} and PhyMask~\citep{kara2024phymask}. LLM-based frameworks integrate sensing into language-centric models~\citep{gruver2023large, garza2023timegpt}, including LIMU-BERT~\citep{xu2021limu}, Penetrative AI~\citep{xu2024penetrative}, LLMSense~\citep{ouyang2024llmsense}, and IoT-LM~\citep{mo2024iot}. These methods share the goal of a transferable representation for IoT sensing but address cross-modality heterogeneity through multi-modal alignment alone, leaving within-modality heterogeneity unmodeled and remaining placement-agnostic with respect to spatial layout. \MODEL{} closes both gaps, instantiated as minimal modifications to the standard MAE pipeline. 

\noindent\textbf{Pretraining with Different Notions of "Spatial" Context.}
Several works incorporate spatial context, though definitions of “spatial” differ. In vision, it typically refers to grids of pixels or patches, as in video~\citep{feichtenhofer2022masked, wu2023dropmae}, remote sensing~\citep{lin2023ss, reed2023scale, irvin2023usat}, and 3D medical imaging~\citep{gu2024self}. Beyond vision, "spatial" often denotes discrete symbolic entities, e.g., joints in SkeletonMAE~\citep{wu2023skeletonmae}, EEG channels in MV-SSTMA~\citep{li2022multi} and MMM~\citep{yi2023learning}, or sensor identities in Gao~\etal~\citep{gao2023spatial} and Miao~\etal~\citep{miao2024spatial}.  These works establish that pretraining benefits from spatial structure when that structure is regular (grids) or discrete (joints, channels, sensor identifiers); \MODEL{} adopts the same intuition---that an encoder should account for where each observation comes from---and generalizes it to the irregular, continuous sensor layouts of distributed sensing, where neither a fixed grid nor a discrete enumeration applies. A related but distinct line is neural scene representation and novel view synthesis~\citep{mildenhall2021nerf,kerbl20233d,wu20244d}: these methods also exploit spatial layouts, but fit a single per-scene model end-to-end on multi-view reconstruction rather than producing a representation transferable to downstream sensing tasks.

\noindent\textbf{Pretraining via Positional Reconstruction Objectives.}
A third line of work, often without using the term “spatial,” incorporates positional reconstruction objectives. In vision, Doersch~\etal~\citep{doersch2015unsupervised} proposed predicting relative patch positions, extended by jigsaw~\citep{noroozi2016unsupervised} and content restoration~\citep{kim2018learning}. DeepPermNet~\citep{santa2017deeppermnet} learns permutation structures, MP3~\citep{zhai2022position} predicts absolute patch locations, and LOCA~\citep{caron2024location} predicts relative positions of clustered patches. In NLP, StructBERT~\citep{wang2019structbert}, ALBERT~\citep{lan2019albert}, and SLM~\citep{lee2020slm} use sentence order prediction and sequence restoration, while Nandy~\etal~\citep{nandy2024order} extend to permutation-based objectives. Beyond vision and language, GeoMAE~\citep{tian2023geomae} reconstructs geometric features of masked point clouds, and LEGO~\citep{sun20243d} recovers perturbed molecular geometries. While conceptually related, these methods operate on discrete, domain-specific positional targets (e.g., patch indices, sentence order). \MODEL{} adopts the same positional-reconstruction idea but applies it to continuous physical sensor positions in distributed sensing.

\section{Discussion and Limitations}
\label{sec:limitations}
The paper strives to make a small step towards an eventual vision of {\em representation learning\/} as a {\em thin waist\/} of intelligent sensor information fusion. We are inspired by the Internet protocol stack, where the Internet Protocol (IP) serves as the thin waist, isolating the heterogeneity below it from that above. Analogously, by offering a concise latent representation of the physical environment, representation learning can decouple, at the encoder level, the exact placement and configuration of distributed sensing sources from the consumers comprising different AI tasks. In an ideal scenario, sensors could be replaced, upgraded, or changed without impact on downstream AI, and different AI analytics could be used without changes to the latent representation or sensors. To achieve this goal, important future work remains.

At present, we use calibration tokens to make the representation robust to sensor configuration. Since hardware, mounting, and environmental factors combine into a signature that is in practice distinct for each sensor, a token that captures them could be naturally identifying; the design asks only that this information be consumed as conditioning rather than retained in the representation. Invariance to sensor heterogeneity remains an aspirational goal; the current evaluation establishes robustness, not invariance. Adding a new sensor type currently requires retraining the encoder, which limits drop-in modularity in evolving deployments; a compartmentalized encoder, in which adding a sensor type has a more local effect (akin to adding a sensor driver), is a promising direction. 

Also, while the goal was to support any unseen future placement, we currently see some degradation when testing a placement not seen during encoder pre-training time. Holding out a site removes both training data and placement diversity from a pretraining set that is modest in both respects, which we believe is the primary reason for the gap; we expect it to be mitigated as more training data becomes available. Likewise, multi-vehicle evaluation is confined to one site and the field study drives vehicles one at a time; broader multi-target evaluation is left to future work. Tolerance to missing data (Table~\ref{tb:m3n-vc4}) is likewise expected to depend on the array's redundancy relative to the task, so the tolerable loss rate is best characterized per deployment rather than by a single threshold. Our evaluation covers static sensor nodes only. We conjecture that the framework may extend to mobile nodes: sensor positions enter as inputs with each sample rather than as fixed model parameters, so a moving array could be treated as a sequence of samples with time-varying layout. This extension is left to future work.

Finally, the approach is geared for omni-directional sensors. It would be interesting to investigate generalizations to directional sensors, although much more work is done on directional data elsewhere (e.g., in the machine vision community), making our emphasis on omni-directional sensing, arguably, a source of novelty.

\section{Conclusion}
\label{sec:conclusion}
This paper presented \MODEL{}, a self-supervised pretraining framework for distributed sensing, as a step towards developing a \emph{thin-waist} of learning-enabled distributed location-aware sensor data processing. \MODEL{} jointly represents signals and the continuous spatial layout of their vantage points, guided by the duality between the two, and uses per-sensor calibration to keep the representation robust despite minor sensor variability. Our experiments show that \MODEL{} (i) does improve the accuracy of downstream spatial analytics, (ii) gracefully handles placements not seen at encoder training time, (iii) benefits from each of its components, (iv) can feasibly run on the edge, and (v) generalizes across several applications.
We hope \MODEL{} will help orient future distributed-sensing representation learning work around improved latent representation of spatial-temporal physical environments and events.

\begin{acks}
The work was supported in part by ARL Cooperative Agreement W911NF-17-2-0196, NSF CNS grant 20-38817, and Boeing Inc.
\end{acks}

\newpage
\bibliographystyle{ACM-Reference-Format}
\bibliography{example_paper}

@inproceedings{ouyang2022cosmo,
  title={Cosmo: contrastive fusion learning with small data for multimodal human activity recognition},
  author={Ouyang, Xiaomin and Shuai, Xian and Zhou, Jiayu and Shi, Ivy Wang and Xie, Zhiyuan and Xing, Guoliang and Huang, Jianwei},
  booktitle={Proceedings of the 28th Annual International Conference on Mobile Computing And Networking},
  pages={324--337},
  year={2022}
}

@article{terven2023comprehensive,
  title={A comprehensive review of yolo architectures in computer vision: From yolov1 to yolov8 and yolo-nas},
  author={Terven, Juan and C{\'o}rdova-Esparza, Diana-Margarita and Romero-Gonz{\'a}lez, Julio-Alejandro},
  journal={Machine learning and knowledge extraction},
  volume={5},
  number={4},
  pages={1680--1716},
  year={2023},
  publisher={mdpi}
}

@article{khan2022transformers,
  title={Transformers in vision: A survey},
  author={Khan, Salman and Naseer, Muzammal and Hayat, Munawar and Zamir, Syed Waqas and Khan, Fahad Shahbaz and Shah, Mubarak},
  journal={ACM computing surveys (CSUR)},
  volume={54},
  number={10s},
  pages={1--41},
  year={2022},
  publisher={ACM New York, NY}
}

@article{zhao2026survey,
  title={A survey of large language models},
  author={Zhao, Wayne Xin and Zhou, Kun and Li, Junyi and Tang, Tianyi and Dong, Zican and Hou, Yupeng and Zhang, Beichen and Min, Yingqian and Zhang, Junjie and Liu, Peiyu and others},
  journal={Frontiers of Computer Science},
  volume={20},
  number={12},
  pages={2012627},
  year={2026},
  publisher={Springer}
}

@article{chen2024survey,
  title={A survey on 3d gaussian splatting},
  author={Chen, Guikun and Wang, Wenguan},
  journal={ACM Computing Surveys},
  year={2024},
  publisher={ACM New York, NY}
}

@article{mildenhall2021nerf,
  title={Nerf: Representing scenes as neural radiance fields for view synthesis},
  author={Mildenhall, Ben and Srinivasan, Pratul P and Tancik, Matthew and Barron, Jonathan T and Ramamoorthi, Ravi and Ng, Ren},
  journal={Communications of the ACM},
  volume={65},
  number={1},
  pages={99--106},
  year={2021},
  publisher={ACM New York, NY, USA}
}

@article{ericsson2022self,
  title={Self-supervised representation learning: Introduction, advances, and challenges},
  author={Ericsson, Linus and Gouk, Henry and Loy, Chen Change and Hospedales, Timothy M},
  journal={IEEE Signal Processing Magazine},
  volume={39},
  number={3},
  pages={42--62},
  year={2022},
  publisher={IEEE}
}

@article{bommasani2021opportunities,
  title={On the opportunities and risks of foundation models},
  author={Bommasani, Rishi and Hudson, Drew A and Adeli, Ehsan and Altman, Russ and Arora, Simran and von Arx, Sydney and Bernstein, Michael S and Bohg, Jeannette and Bosselut, Antoine and Brunskill, Emma and others},
  journal={arXiv preprint arXiv:2108.07258},
  year={2021}
}

@article{deldari2022cocoa,
  title={Cocoa: Cross modality contrastive learning for sensor data},
  author={Deldari, Shohreh and Xue, Hao and Saeed, Aaqib and Smith, Daniel V and Salim, Flora D},
  journal={Proceedings of the ACM on Interactive, Mobile, Wearable and Ubiquitous Technologies},
  volume={6},
  number={3},
  pages={1--28},
  year={2022},
  publisher={ACM New York, NY, USA}
}

@article{liu2023focal,
  title={Focal: Contrastive learning for multimodal time-series sensing signals in factorized orthogonal latent space},
  author={Liu, Shengzhong and Kimura, Tomoyoshi and Liu, Dongxin and Wang, Ruijie and Li, Jinyang and Diggavi, Suhas and Srivastava, Mani and Abdelzaher, Tarek},
  journal={Advances in Neural Information Processing Systems},
  volume={36},
  pages={47309--47338},
  year={2023}
}

@inproceedings{girdhar2023imagebind,
  title={Imagebind: One embedding space to bind them all},
  author={Girdhar, Rohit and El-Nouby, Alaaeldin and Liu, Zhuang and Singh, Mannat and Alwala, Kalyan Vasudev and Joulin, Armand and Misra, Ishan},
  booktitle={Proceedings of the IEEE/CVF conference on computer vision and pattern recognition},
  pages={15180--15190},
  year={2023}
}

@article{ouyang2024mmbind,
  title={MMBind: Unleashing the Potential of Distributed and Heterogeneous Data for Multimodal Learning in IoT},
  author={Ouyang, Xiaomin and Wu, Jason and Kimura, Tomoyoshi and Lin, Yihan and Verma, Gunjan and Abdelzaher, Tarek and Srivastava, Mani},
  journal={arXiv preprint arXiv:2411.12126},
  year={2024}
}

@article{li2023ti,
  title={Ti-mae: Self-supervised masked time series autoencoders},
  author={Li, Zhe and Rao, Zhongwen and Pan, Lujia and Wang, Pengyun and Xu, Zenglin},
  journal={arXiv preprint arXiv:2301.08871},
  year={2023}
}

@article{goswami2024moment,
  title={Moment: A family of open time-series foundation models},
  author={Goswami, Mononito and Szafer, Konrad and Choudhry, Arjun and Cai, Yifu and Li, Shuo and Dubrawski, Artur},
  journal={arXiv preprint arXiv:2402.03885},
  year={2024}
}

@article{liu2025ts,
  title={TS-MAE: A masked autoencoder for time series representation learning},
  author={Liu, Qian and Ye, Junchen and Liang, Haohan and Sun, Leilei and Du, Bowen},
  journal={Information Sciences},
  volume={690},
  pages={121576},
  year={2025},
  publisher={Elsevier}
}

@inproceedings{kara2024freqmae,
  title={Freqmae: Frequency-aware masked autoencoder for multi-modal iot sensing},
  author={Kara, Denizhan and Kimura, Tomoyoshi and Liu, Shengzhong and Li, Jinyang and Liu, Dongxin and Wang, Tianshi and Wang, Ruijie and Chen, Yizhuo and Hu, Yigong and Abdelzaher, Tarek},
  booktitle={Proceedings of the ACM Web Conference 2024},
  pages={2795--2806},
  year={2024}
}

@inproceedings{kara2024phymask,
  title={PhyMask: An Adaptive Masking Paradigm for Efficient Self-Supervised Learning in IoT},
  author={Kara, Denizhan and Kimura, Tomoyoshi and Chen, Yatong and Li, Jinyang and Wang, Ruijie and Chen, Yizhuo and Wang, Tianshi and Liu, Shengzhong and Abdelzaher, Tarek},
  booktitle={Proceedings of the 22nd ACM Conference on Embedded Networked Sensor Systems},
  pages={97--111},
  year={2024}
}

@inproceedings{xu2021limu,
  title={Limu-bert: Unleashing the potential of unlabeled data for imu sensing applications},
  author={Xu, Huatao and Zhou, Pengfei and Tan, Rui and Li, Mo and Shen, Guobin},
  booktitle={Proceedings of the 19th ACM Conference on Embedded Networked Sensor Systems},
  pages={220--233},
  year={2021}
}

@inproceedings{ouyang2024llmsense,
  title={LLMSense: Harnessing LLMs for high-level reasoning over spatiotemporal sensor traces},
  author={Ouyang, Xiaomin and Srivastava, Mani},
  booktitle={2024 IEEE 3rd Workshop on Machine Learning on Edge in Sensor Systems (SenSys-ML)},
  pages={9--14},
  year={2024},
  organization={IEEE}
}

@inproceedings{xu2024penetrative,
  title={Penetrative ai: Making llms comprehend the physical world},
  author={Xu, Huatao and Han, Liying and Yang, Qirui and Li, Mo and Srivastava, Mani},
  booktitle={Proceedings of the 25th International Workshop on Mobile Computing Systems and Applications},
  pages={1--7},
  year={2024}
}

@article{mo2024iot,
  title={Iot-lm: Large multisensory language models for the internet of things},
  author={Mo, Shentong and Salakhutdinov, Russ and Morency, Louis-Philippe and Liang, Paul Pu},
  journal={arXiv preprint arXiv:2407.09801},
  year={2024}
}

@article{woo2024unified,
  title={Unified training of universal time series forecasting transformers},
  author={Woo, Gerald and Liu, Chenghao and Kumar, Akshat and Xiong, Caiming and Savarese, Silvio and Sahoo, Doyen},
  year={2024},
  publisher={PMLR}
}

@inproceedings{das2024decoder,
  title={A decoder-only foundation model for time-series forecasting},
  author={Das, Abhimanyu and Kong, Weihao and Sen, Rajat and Zhou, Yichen},
  booktitle={Forty-first International Conference on Machine Learning},
  year={2024}
}

@article{gruver2023large,
  title={Large language models are zero-shot time series forecasters},
  author={Gruver, Nate and Finzi, Marc and Qiu, Shikai and Wilson, Andrew G},
  journal={Advances in Neural Information Processing Systems},
  volume={36},
  pages={19622--19635},
  year={2023}
}

@article{garza2023timegpt,
  title={TimeGPT-1},
  author={Garza, Azul and Challu, Cristian and Mergenthaler-Canseco, Max},
  journal={arXiv preprint arXiv:2310.03589},
  year={2023}
}

@inproceedings{tian2020contrastive,
  title={Contrastive multiview coding},
  author={Tian, Yonglong and Krishnan, Dilip and Isola, Phillip},
  booktitle={Computer Vision--ECCV 2020: 16th European Conference, Glasgow, UK, August 23--28, 2020, Proceedings, Part XI 16},
  pages={776--794},
  year={2020},
  organization={Springer}
}

@inproceedings{chen2020simple,
  title={A simple framework for contrastive learning of visual representations},
  author={Chen, Ting and Kornblith, Simon and Norouzi, Mohammad and Hinton, Geoffrey},
  booktitle={International conference on machine learning},
  pages={1597--1607},
  year={2020},
  organization={PmLR}
}

@article{huang2022masked,
  title={Masked autoencoders that listen},
  author={Huang, Po-Yao and Xu, Hu and Li, Juncheng and Baevski, Alexei and Auli, Michael and Galuba, Wojciech and Metze, Florian and Feichtenhofer, Christoph},
  journal={Advances in Neural Information Processing Systems},
  volume={35},
  pages={28708--28720},
  year={2022}
}

@article{gong2022contrastive,
  title={Contrastive audio-visual masked autoencoder},
  author={Gong, Yuan and Rouditchenko, Andrew and Liu, Alexander H and Harwath, David and Karlinsky, Leonid and Kuehne, Hilde and Glass, James},
  journal={arXiv preprint arXiv:2210.07839},
  year={2022}
}

@inproceedings{noroozi2016unsupervised,
  title={Unsupervised learning of visual representations by solving jigsaw puzzles},
  author={Noroozi, Mehdi and Favaro, Paolo},
  booktitle={European conference on computer vision},
  pages={69--84},
  year={2016},
  organization={Springer}
}

@inproceedings{doersch2015unsupervised,
  title={Unsupervised visual representation learning by context prediction},
  author={Doersch, Carl and Gupta, Abhinav and Efros, Alexei A},
  booktitle={Proceedings of the IEEE international conference on computer vision},
  pages={1422--1430},
  year={2015}
}

@inproceedings{santa2017deeppermnet,
  title={Deeppermnet: Visual permutation learning},
  author={Santa Cruz, Rodrigo and Fernando, Basura and Cherian, Anoop and Gould, Stephen},
  booktitle={Proceedings of the IEEE Conference on Computer Vision and Pattern Recognition},
  pages={3949--3957},
  year={2017}
}

@article{zhai2022position,
  title={Position prediction as an effective pretraining strategy},
  author={Zhai, Shuangfei and Jaitly, Navdeep and Ramapuram, Jason and Busbridge, Dan and Likhomanenko, Tatiana and Cheng, Joseph Yitan and Talbott, Walter and Huang, Chen and Goh, Hanlin and Susskind, Joshua},
  journal={arXiv preprint arXiv:2207.07611},
  year={2022}
}

@inproceedings{kim2018learning,
  title={Learning image representations by completing damaged jigsaw puzzles},
  author={Kim, Dahun and Cho, Donghyeon and Yoo, Donggeun and Kweon, In So},
  booktitle={2018 IEEE winter conference on applications of computer vision (WACV)},
  pages={793--802},
  year={2018},
  organization={IEEE}
}

@article{wang2019structbert,
  title={StructBERT: Incorporating Language Structures into Pre-training for Deep Language Understanding. eprint},
  author={Wang, W and Bi, B and Yan, M and Wu, C and Bao, Z and Xia, J},
  journal={arXiv preprint arXiv:1908.04577},
  year={2019}
}

@article{nandy2024order,
  title={Order-Based Pre-training Strategies for Procedural Text Understanding},
  author={Nandy, Abhilash and Kulkarni, Yash and Goyal, Pawan and Ganguly, Niloy},
  journal={arXiv preprint arXiv:2404.04676},
  year={2024}
}

@article{lee2020slm,
  title={SLM: Learning a discourse language representation with sentence unshuffling},
  author={Lee, Haejun and Hudson, Drew A and Lee, Kangwook and Manning, Christopher D},
  journal={arXiv preprint arXiv:2010.16249},
  year={2020}
}

@article{lan2019albert,
  title={Albert: A lite bert for self-supervised learning of language representations},
  author={Lan, Zhenzhong and Chen, Mingda and Goodman, Sebastian and Gimpel, Kevin and Sharma, Piyush and Soricut, Radu},
  journal={arXiv preprint arXiv:1909.11942},
  year={2019}
}

@inproceedings{tian2023geomae,
  title={Geomae: Masked geometric target prediction for self-supervised point cloud pre-training},
  author={Tian, Xiaoyu and Ran, Haoxi and Wang, Yue and Zhao, Hang},
  booktitle={Proceedings of the IEEE/CVF Conference on Computer Vision and Pattern Recognition},
  pages={13570--13580},
  year={2023}
}

@article{sun20243d,
  title={3D Molecular Pretraining via Localized Geometric Generation},
  author={Sun, Yuancheng and Chen, Kai and Liu, Kang and Ye, Qiwei},
  journal={bioRxiv},
  pages={2024--09},
  year={2024},
  publisher={Cold Spring Harbor Laboratory}
}

@inproceedings{caron2024location,
  title={Location-aware self-supervised transformers for semantic segmentation},
  author={Caron, Mathilde and Houlsby, Neil and Schmid, Cordelia},
  booktitle={Proceedings of the IEEE/CVF Winter Conference on Applications of Computer Vision},
  pages={117--127},
  year={2024}
}

@inproceedings{kong2023understanding,
  title={Understanding masked image modeling via learning occlusion invariant feature},
  author={Kong, Xiangwen and Zhang, Xiangyu},
  booktitle={Proceedings of the IEEE/CVF Conference on Computer Vision and Pattern Recognition},
  pages={6241--6251},
  year={2023}
}

@inproceedings{li2022exploring,
  title={Exploring plain vision transformer backbones for object detection},
  author={Li, Yanghao and Mao, Hanzi and Girshick, Ross and He, Kaiming},
  booktitle={European conference on computer vision},
  pages={280--296},
  year={2022},
  organization={Springer}
}

@article{feichtenhofer2022masked,
  title={Masked autoencoders as spatiotemporal learners},
  author={Feichtenhofer, Christoph and Li, Yanghao and He, Kaiming and others},
  journal={Advances in neural information processing systems},
  volume={35},
  pages={35946--35958},
  year={2022}
}

@inproceedings{wu2023skeletonmae,
  title={Skeletonmae: Spatial-temporal masked autoencoders for self-supervised skeleton action recognition},
  author={Wu, Wenhan and Hua, Yilei and Zheng, Ce and Wu, Shiqian and Chen, Chen and Lu, Aidong},
  booktitle={2023 IEEE international conference on multimedia and expo workshops (ICMEW)},
  pages={224--229},
  year={2023},
  organization={IEEE}
}

@inproceedings{reed2023scale,
  title={Scale-mae: A scale-aware masked autoencoder for multiscale geospatial representation learning},
  author={Reed, Colorado J and Gupta, Ritwik and Li, Shufan and Brockman, Sarah and Funk, Christopher and Clipp, Brian and Keutzer, Kurt and Candido, Salvatore and Uyttendaele, Matt and Darrell, Trevor},
  booktitle={Proceedings of the IEEE/CVF International Conference on Computer Vision},
  pages={4088--4099},
  year={2023}
}

@inproceedings{wu2023dropmae,
  title={Dropmae: Masked autoencoders with spatial-attention dropout for tracking tasks},
  author={Wu, Qiangqiang and Yang, Tianyu and Liu, Ziquan and Wu, Baoyuan and Shan, Ying and Chan, Antoni B},
  booktitle={Proceedings of the IEEE/CVF conference on computer vision and pattern recognition},
  pages={14561--14571},
  year={2023}
}

@article{lin2023ss,
  title={SS-MAE: Spatial--spectral masked autoencoder for multisource remote sensing image classification},
  author={Lin, Junyan and Gao, Feng and Shi, Xiaochen and Dong, Junyu and Du, Qian},
  journal={IEEE Transactions on Geoscience and Remote Sensing},
  volume={61},
  pages={1--14},
  year={2023},
  publisher={IEEE}
}

@article{gao2023spatial,
  title={Spatial-temporal-decoupled masked pre-training for spatiotemporal forecasting},
  author={Gao, Haotian and Jiang, Renhe and Dong, Zheng and Deng, Jinliang and Ma, Yuxin and Song, Xuan},
  journal={arXiv preprint arXiv:2312.00516},
  year={2023}
}

@article{irvin2023usat,
  title={USat: A unified self-supervised encoder for multi-sensor satellite imagery},
  author={Irvin, Jeremy and Tao, Lucas and Zhou, Joanne and Ma, Yuntao and Nashold, Langston and Liu, Benjamin and Ng, Andrew Y},
  journal={arXiv preprint arXiv:2312.02199},
  year={2023}
}

@article{gu2024self,
  title={Self pre-training with topology-and spatiality-aware masked autoencoders for 3D medical image segmentation},
  author={Gu, Pengfei and Zhang, Yejia and Li, Huimin and Wang, Chaoli and Chen, Danny Z},
  journal={arXiv preprint arXiv:2406.10519},
  year={2024}
}

@inproceedings{li2022multi,
  title={A multi-view spectral-spatial-temporal masked autoencoder for decoding emotions with self-supervised learning},
  author={Li, Rui and Wang, Yiting and Zheng, Wei-Long and Lu, Bao-Liang},
  booktitle={Proceedings of the 30th ACM International Conference on Multimedia},
  pages={6--14},
  year={2022}
}

@article{miao2024spatial,
  title={Spatial-temporal masked autoencoder for multi-device wearable human activity recognition},
  author={Miao, Shenghuan and Chen, Ling and Hu, Rong},
  journal={Proceedings of the ACM on Interactive, Mobile, Wearable and Ubiquitous Technologies},
  volume={7},
  number={4},
  pages={1--25},
  year={2024},
  publisher={ACM New York, NY, USA}
}

@article{yi2023learning,
  title={Learning topology-agnostic eeg representations with geometry-aware modeling},
  author={Yi, Ke and Wang, Yansen and Ren, Kan and Li, Dongsheng},
  journal={Advances in Neural Information Processing Systems},
  volume={36},
  pages={53875--53891},
  year={2023}
}

@article{si2024all,
  title={An all-in-one seismic phase picking, location, and association network for multi-task multi-station earthquake monitoring},
  author={Si, Xu and Wu, Xinming and Li, Zefeng and Wang, Shenghou and Zhu, Jun},
  journal={Communications Earth \& Environment},
  volume={5},
  number={1},
  pages={22},
  year={2024},
  publisher={Nature Publishing Group UK London}
}

@article{center2013southern,
  title={Southern California Earthquake Center},
  author={Center, Earthquake},
  journal={Caltech. Dataset},
  volume={394},
  year={2013}
}

@misc{caltech1926socal,
  author       = {{California Institute of Technology (Caltech)}},
  title        = {Southern California Seismic Network},
  year         = {1926},
  publisher    = {International Federation of Digital Seismograph Networks},
  howpublished = {Other/Seismic Network},
  doi          = {10.7914/SN/CI}
}

@inproceedings{li2025restoreml,
  author    = {Jinyang Li and Yizhuo Chen and Ruijie Wang and Tomoyoshi Kimura and Tianshi Wang and You Lyu and Hongjue Zhao and Binqi Sun and Shangchen Wu and Yigong Hu and Denizhan Kara and Beitong Tian and Klara Nahrstedt and Suhas Diggavi and Jae H. Kim and Greg Kimberly and Guijun Wang and Maggie Wigness and Tarek Abdelzaher},
  title     = {{RestoreML}: Practical Unsupervised Tuning of Deployed Intelligent IoT Systems},
  booktitle = {2025 The 21st International Conference on Distributed Computing in Smart Systems and the Internet of Things (DCOSS-IoT)},
  year      = {2025},
  publisher = {IEEE}
}

@inproceedings{sztyler2016body,
  title={On-body localization of wearable devices: An investigation of position-aware activity recognition},
  author={Sztyler, Timo and Stuckenschmidt, Heiner},
  booktitle={2016 IEEE international conference on pervasive computing and communications (PerCom)},
  pages={1--9},
  year={2016},
  organization={IEEE}
}

@book{united1987department,
  title={Department of Defense World Geodetic System 1984: its definition and relationships with local geodetic systems},
  author={United States. Defense Mapping Agency},
  volume={8350},
  year={1987},
  publisher={Defense Mapping Agency}
}

@article{gu2021survey,
  title={A survey on deep learning for human activity recognition},
  author={Gu, Fuqiang and Chung, Mu-Huan and Chignell, Mark and Valaee, Shahrokh and Zhou, Baoding and Liu, Xue},
  journal={ACM Computing Surveys (CSUR)},
  volume={54},
  number={8},
  pages={1--34},
  year={2021},
  publisher={ACM New York, NY}
}

@article{syed2021iot,
  title={IoT in smart cities: A survey of technologies, practices and challenges},
  author={Syed, Abbas Shah and Sierra-Sosa, Daniel and Kumar, Anup and Elmaghraby, Adel},
  journal={Smart Cities},
  volume={4},
  number={2},
  pages={429--475},
  year={2021},
  publisher={MDPI}
}

@article{bathla2022autonomous,
  title={Autonomous vehicles and intelligent automation: Applications, challenges, and opportunities},
  author={Bathla, Gourav and Bhadane, Kishor and Singh, Rahul Kumar and Kumar, Rajneesh and Aluvalu, Rajanikanth and Krishnamurthi, Rajalakshmi and Kumar, Adarsh and Thakur, RN and Basheer, Shakila},
  journal={Mobile Information Systems},
  volume={2022},
  number={1},
  pages={7632892},
  year={2022},
  publisher={Wiley Online Library}
}

@article{ullo2020advances,
  title={Advances in smart environment monitoring systems using IoT and sensors},
  author={Ullo, Silvia Liberata and Sinha, Ganesh Ram},
  journal={Sensors},
  volume={20},
  number={11},
  pages={3113},
  year={2020},
  publisher={MDPI}
}

@inproceedings{he2022masked,
  title={Masked autoencoders are scalable vision learners},
  author={He, Kaiming and Chen, Xinlei and Xie, Saining and Li, Yanghao and Doll{\'a}r, Piotr and Girshick, Ross},
  booktitle={Proceedings of the IEEE/CVF conference on computer vision and pattern recognition},
  pages={16000--16009},
  year={2022}
}

@article{kerbl20233d,
  title={3d gaussian splatting for real-time radiance field rendering.},
  author={Kerbl, Bernhard and Kopanas, Georgios and Leimk{\"u}hler, Thomas and Drettakis, George},
  journal={ACM Trans. Graph.},
  volume={42},
  number={4},
  pages={139--1},
  year={2023}
}

@inproceedings{wu20244d,
  title={4d gaussian splatting for real-time dynamic scene rendering},
  author={Wu, Guanjun and Yi, Taoran and Fang, Jiemin and Xie, Lingxi and Zhang, Xiaopeng and Wei, Wei and Liu, Wenyu and Tian, Qi and Wang, Xinggang},
  booktitle={Proceedings of the IEEE/CVF conference on computer vision and pattern recognition},
  pages={20310--20320},
  year={2024}
}

@article{dosovitskiy2020image,
  title={An image is worth 16x16 words: Transformers for image recognition at scale},
  author={Dosovitskiy, Alexey and Beyer, Lucas and Kolesnikov, Alexander and Weissenborn, Dirk and Zhai, Xiaohua and Unterthiner, Thomas and Dehghani, Mostafa and Minderer, Matthias and Heigold, Georg and Gelly, Sylvain and others},
  journal={arXiv preprint arXiv:2010.11929},
  year={2020}
}

@article{vaswani2017attention,
  title={Attention is all you need},
  author={Vaswani, Ashish and Shazeer, Noam and Parmar, Niki and Uszkoreit, Jakob and Jones, Llion and Gomez, Aidan N and Kaiser, {\L}ukasz and Polosukhin, Illia},
  journal={Advances in neural information processing systems},
  volume={30},
  year={2017}
}

@article{su2024roformer,
  title={Roformer: Enhanced transformer with rotary position embedding},
  author={Su, Jianlin and Ahmed, Murtadha and Lu, Yu and Pan, Shengfeng and Bo, Wen and Liu, Yunfeng},
  journal={Neurocomputing},
  volume={568},
  pages={127063},
  year={2024},
  publisher={Elsevier}
}

@article{press2021train,
  title={Train short, test long: Attention with linear biases enables input length extrapolation},
  author={Press, Ofir and Smith, Noah A and Lewis, Mike},
  journal={arXiv preprint arXiv:2108.12409},
  year={2021}
}

@inproceedings{carion2020end,
  title={End-to-end object detection with transformers},
  author={Carion, Nicolas and Massa, Francisco and Synnaeve, Gabriel and Usunier, Nicolas and Kirillov, Alexander and Zagoruyko, Sergey},
  booktitle={European conference on computer vision},
  pages={213--229},
  year={2020},
  organization={Springer}
}

@inproceedings{sohn2015cvae,
  title={Learning structured output representation using deep conditional generative models},
  author={Sohn, Kihyuk and Lee, Honglak and Yan, Xinchen},
  booktitle={Advances in Neural Information Processing Systems},
  year={2015}
}

@inproceedings{lample2017fader,
  title={Fader Networks: Manipulating Images by Sliding Attributes},
  author={Lample, Guillaume and Zeghidour, Neil and Usunier, Nicolas and Bordes, Antoine and Denoyer, Ludovic and Ranzato, Marc'Aurelio},
  booktitle={Advances in Neural Information Processing Systems},
  year={2017}
}

@inproceedings{bouchacourt2018mlvae,
  title={Multi-level variational autoencoder: Learning disentangled representations from grouped observations},
  author={Bouchacourt, Diane and Tomioka, Ryota and Nowozin, Sebastian},
  booktitle={AAAI Conference on Artificial Intelligence},
  year={2018}
}

@article{shwartzziv2017opening,
  title={Opening the black box of deep neural networks via information},
  author={Shwartz-Ziv, Ravid and Tishby, Naftali},
  journal={arXiv preprint arXiv:1703.00810},
  year={2017}
}

@article{achille2018emergence,
  title={Emergence of invariance and disentanglement in deep representations},
  author={Achille, Alessandro and Soatto, Stefano},
  journal={Journal of Machine Learning Research},
  volume={19},
  number={50},
  pages={1--34},
  year={2018}
}

@article{zhang2025multistation,
  title={Multistation seismic location via machine learning: application to Oklahoma and Southern California},
  author={Zhang, Ji and Zhu, Huiyu and Li, Zefeng and Wu, Xinming and Zhang, Jie},
  journal={Geophysical Journal International},
  volume={241},
  number={3},
  pages={1853--1867},
  year={2025},
  publisher={Oxford University Press}
}

@article{munchmeyer2021earthquake,
  title={Earthquake magnitude and location estimation from real time seismic waveforms with a transformer network},
  author={M{\"u}nchmeyer, Jannes and Bindi, Dino and Leser, Ulf and Tilmann, Frederik},
  journal={Geophysical Journal International},
  volume={226},
  number={2},
  pages={1086--1104},
  year={2021},
  publisher={Oxford University Press}
}

\appendix

\section*{Appendix 1: Proof of Proposition \ref{pp:information}}
\begin{proof}\label{pf:information}
In this proof, we use $C$ and $C'$ to denote generic constants independent of model parameters, whose specific values may change from Equation to Equation. For brevity, the proof is stated for fully observed samples; with missing entries, the same argument applies unchanged to the observed entries, which are the only ones that enter the loss.

\noindent\textbf{Classical MAE.} We begin with the case of classical MAE. We assume, following prior work~\citep{li2022exploring,kong2023understanding}, that due to the high dimension of the encoder output $\mZ$ relative to the original signal, $\mZ$ retains the full information about the unmasked part of each modality's signals $\mask({\tX^{(k)};\mM^{(k)}})$, which the corresponding per-modality decoder can reconstruct with negligible loss. As a result, we consider the reconstruction loss calculated on the masked signals to be equivalent to the reconstruction loss calculated on the full signals:
\begin{equation}
\begin{aligned}
    L' &= \sum_{k=1}^K \| \mask(\tX^{(k)}-\widehat{\tX}^{(k)};\overline{\mM}^{(k)})\|_2^2 \\
    &= \sum_{k=1}^K  \| \tX^{(k)}-\widehat{\tX}^{(k)}\|_2^2 \\
    & = \sum_{k=1}^K L'^{(k)},
\end{aligned}
\end{equation}
where $L'^{(k)}$ is the reconstruction loss calculated for the $k$-th modality.

Like in the analysis for general regression tasks, the likelihood $P_\mathrm{dec}(\mathsf{X}^{(k)} \mid \mathsf{Z}=\mZ)$ implicitly modeled by the decoder is defined as a fully factorized Gaussian distribution with mean $\widehat{\tX}^{(k)}$:
\begin{equation}
    P_\mathrm{dec}(\mathsf{X}^{(k)} \mid \mathsf{Z}=\mZ) \overset{\underset{\mathrm{def}}{}}{=} \gN(\widehat{\tX}^{(k)}, \frac{1}{\sqrt{2}}\mathrm{I}).
\end{equation}

Then, we can interpret the MSE loss $L'^{(k)}$ as proportional to the negative log-likelihood:
\begin{equation}\label{eq:nll}
\begin{aligned}
    -\log P_\mathrm{dec}(\mathsf{X}^{(k)}=\tX^{(k)} \mid \mathsf{Z}=\mZ) & = \| \tX^{(k)}-\widehat{\tX}^{(k)}\|_2^2 + C' \\
    &= L'^{(k)} + C'.
\end{aligned}
\end{equation}

Since the prior probability $P(\mathsf{X}^{(k)}=\tX^{(k)})$ is also a constant independent of the model parameters (only determined by the dataset distribution), we can further have
\begin{equation}
\begin{aligned}
    L'^{(k)} + C' = -\log  \frac{ P_\mathrm{dec}(\mathsf{X}^{(k)} =\tX^{(k)} \mid \mathsf{Z}=\mZ)}{P(\mathsf{X}^{(k)} =\tX^{(k)})}.
\end{aligned}
\end{equation}

Taking expectation over the data distribution and applying the standard mutual information decomposition, we can have:
\begin{equation}\label{eq:bound}
\begin{aligned}
    \E[L'^{(k)}] + C'
    & = \E\left[-\log  \frac{ P_\mathrm{dec}(\mathsf{X}^{(k)} \mid \mathsf{Z})}{P(\mathsf{X}^{(k)} )}\right] \\
    & = \E\bigg[-\log  \frac{ P(\mathsf{X}^{(k)} \mid \mathsf{Z})}{P(\mathsf{X}^{(k)} )} \\
    & \qquad + \log  \frac{ P(\mathsf{X}^{(k)} \mid \mathsf{Z})}{P_\mathrm{dec}(\mathsf{X}^{(k)} \mid \mathsf{Z})}\bigg] \\
    & = - I(\mathsf{X}^{(k)}; \mathsf{Z}) \\
    & \quad + KL(P(\mathsf{X}^{(k)} \mid \mathsf{Z}) \,\|\, P_\mathrm{dec}(\mathsf{X}^{(k)} \mid \mathsf{Z})) \\
    & \geq - I(\mathsf{X}^{(k)}; \mathsf{Z}),
\end{aligned}
\end{equation}
where $P(\mathsf{X}^{(k)} \mid \mathsf{Z})$ denotes the non-tractable ground-truth conditional distribution determined by the data distribution and the encoders, and $KL(\cdot \,\|\, \cdot)$ denotes Kullback--Leibler divergence.

Summing over all modalities, we can prove:
\begin{equation}
\begin{aligned}
    -\E[L'] + C' \leq  \sum_{k=1}^K I(\mathsf{X}^{(k)}; \mathsf{Z}).
\end{aligned}
\end{equation}

\bigskip

\noindent\textbf{\MODEL{}.}
For \MODEL{}, the signal decoder takes additional inputs: masked spatial positional and per-sensor calibration embeddings, and consumes the joint representation $\mZ$ (rather than a per-modality encoder output). Let $\mathsf{S}_\mM^{(k)}$ and $\mathsf{R}_\mM^{(k)}$ denote the masked spatial positions and per-sensor calibration tokens, respectively. Let $L_\text{sig}^{(k)}$ denote the signal reconstruction loss for modality $k$. Then, adjusting our reasoning above, we can modify Equation~\ref{eq:nll} to:
\begin{equation}
\begin{aligned}
    L_\mathrm{sig}^{(k)} + C
    & = -\log P_\mathrm{dec}(\mathsf{X}^{(k)}=\tX^{(k)} \mid \mathsf{Z}=\mZ, \\
    & \qquad\quad\, \mathsf{S}_\mM^{(k)}=\tS_\mM^{(k)}, \mathsf{R}_\mM^{(k)}=\tR_\mM^{(k)}).
\end{aligned}
\end{equation}

In \MODEL{}, the latent representation $\mZ$ is computed not only from the unmasked signals but also from the unmasked spatial positional embeddings (across all modalities), so the assumption above extends to them: $\mZ$ retains their full information. The per-sensor calibration tokens need no such assumption, because the decoder receives them directly as input. As a result, we can equivalently condition the likelihood on full spatial positions and per-sensor calibration tokens:
\begin{equation}
\begin{aligned}
    L_\mathrm{sig}^{(k)} + C
    & = -\log P_\mathrm{dec}(\mathsf{X}^{(k)}=\tX^{(k)} \mid \mathsf{Z}=\mZ, \\
    & \qquad\quad\, \mathsf{S}_\mM^{(k)}=\tS_\mM^{(k)}, \mathsf{R}_\mM^{(k)}=\tR_\mM^{(k)}) \\
    & = -\log P_\mathrm{dec}(\mathsf{X}^{(k)}=\tX^{(k)} \mid \mathsf{Z}=\mZ, \\
    & \qquad\quad\, \mathsf{S}^{(k)}=\tS^{(k)}, \mathsf{R}^{(k)}=\tR^{(k)}).
\end{aligned}
\end{equation}

As in the reasoning above, since the prior probability $P(\mathsf{X}^{(k)}=\tX^{(k)}\mid\mathsf{S}^{(k)}=\tS^{(k)}, \mathsf{R}^{(k)}=\tR^{(k)})$ is also independent of the model parameters, we can adjust Equation~\ref{eq:bound} to:
\begin{equation}
\begin{aligned}
    \E[L_\mathrm{sig}^{(k)}] + C \geq - I(\mathsf{X}^{(k)}; \mathsf{Z} \mid \mathsf{S}^{(k)}, \mathsf{R}^{(k)}).
\end{aligned}
\end{equation}

Since \MODEL{} treats spatial positions symmetrically to signals, we can apply the same reasoning to the spatial reconstruction loss $L_\mathrm{sp}^{(k)}$, yielding:
\begin{equation}
\begin{aligned}
    \E[L_\mathrm{sp}^{(k)}] + C \geq - I(\mathsf{S}^{(k)}; \mathsf{Z} \mid \mathsf{X}^{(k)}, \mathsf{R}^{(k)}).
\end{aligned}
\end{equation}

Summing over all modalities and both reconstruction losses, we can prove:
\begin{equation}
\begin{aligned}
    -\E[L] + C
    \leq  \sum_{k=1}^K & I(\mathsf{X}^{(k)}; \mathsf{Z} \mid \mathsf{S}^{(k)}, \mathsf{R}^{(k)}) \\
    & + I(\mathsf{S}^{(k)}; \mathsf{Z} \mid \mathsf{X}^{(k)}, \mathsf{R}^{(k)}).
\end{aligned}
\end{equation}

\end{proof}

\section*{Appendix 2: M3N-VC Site Descriptions}\label{pp:sites}

Across the evaluation, we refer to specific M3N-VC sites by their identifiers (e.g., H08, H24, I22). Table~\ref{table:datasets} reproduces the per-site characteristics from the original M3N-VC release~\citep{li2025restoreml}. Collectively, the six sites span four physically different areas (represented by the first letter of their IDs), each with its own terrain type (asphalt, gravel mixtures, dirt, and concrete). Sites within an area do not necessarily share a sensor layout: H08 and H24 use similar node configurations, whereas I22 and I29 use different placements, with arrays spanning approximately $22$\,m\,$\times$\,$67$\,m and $38$\,m\,$\times$\,$159$\,m (east--west\,$\times$\,north--south), respectively. The six sites also span three weather conditions (sunny, rainy, and windy), four vehicle makes and models, and two sensor-array sizes (6 and 8 nodes). 
\begin{table}[h]
   \centering
   \caption{Per-site characteristics of the M3N-VC dataset, reproduced from Li~\etal~\citep{li2025restoreml}. Vehicle abbreviations: C (Mazda CX-30), G (Mercedes-Benz GLE-350), M (Ford Mustang), X (Mazda MX-5).}
   \label{table:datasets}
   \resizebox{0.98\columnwidth}{!}{%
   \begin{tabular}{ccccccc}
       \toprule
       Site ID &  Terrain & Weather & Targets & \# Nodes & Length  \\
       \midrule
       H08 &  Asphalt \& gravel & Sunny & C, G, M, X & 6 & 2.77 h \\
       H24 &  Asphalt \& gravel & Rainy & C, G, M, X & 6 & 3.43 h \\
       S31 &  Dirt \& gravel & Sunny & C, G, M, X & 6 & 2.78 h \\
       A06 &  Asphalt & Sunny & C, G, X & 6 & 2.14 h  \\
       I29 & Concrete & Windy & C, G, M, X & 8 & 4.14 h  \\
       I22 &  Concrete & Sunny & C, M, X & 8 & 3.00 h  \\
       \bottomrule
   \end{tabular}}
\end{table}

\clearpage
\section*{Appendix 3: Notation}
Table~\ref{tb:notation} summarizes the notation used in Section~\ref{sec:method}.
\begin{table}[H]
\caption{Summary of the notations and their corresponding dimensions and definitions.}
\label{tb:notation}
\begin{center}
\scriptsize
\begin{tabular}{c|c|c}
\toprule
Notation                        &  Dimension(s)                                         &  Definition  \\\midrule
$K$                             &  $\sN$                                                & Number of modalities \\
$n^{(k)}$ &  $\sN$                                                & Number of sensors \\
$m^{(k)}$ &  $\sN$                                                & Number of tokens \\
$m_\mM^{(k)}$ &  $\sN$                                                & Number of visible tokens \\
$d, d_\tX^{(k)}$                &  $\sN$                                                & Model dimension, tokenized signal dimension \\
$d_\mS$                  &  $\sN$                                                & Spatial-position dimension  \\
$d_\mR$                  &  $\sN$                                                & Per-sensor-calibration-token dimension  \\
$L, L'$                         & $\sN$                                                 & Loss function of \MODEL{} and classical MAE \\
$\tX^{(k)}$                     & $\sR^{n^{(k)} \times m^{(k)} \times d_\tX^{(k)}}$     & Signals \\
$\widehat{\tX}^{(k)}$           & $\sR^{n^{(k)} \times m^{(k)} \times d_\tX^{(k)}}$     & Reconstructed  signals \\
$\mathsf{X}^{(k)}$              & $\sR^{n^{(k)} \times m^{(k)} \times d_\tX^{(k)}}$     & Signals random variable \\
$\widetilde{\tX}^{(k)}$         & $\sR^{n^{(k)} \times m^{(k)} \times d}$               & Signal embeddings \\
$\mS^{(k)}$                     &  $\sR^{n^{(k)} \times d_\mS}$                         & Spatial positions \\
$\tS^{(k)}$                     & $\sR^{n^{(k)} \times m^{(k)} \times d_\mS}$           & Spatial positions (broadcasted) \\
$\widehat{\tS}^{(k)}$           & $\sR^{n^{(k)} \times m^{(k)} \times d_\mS}$           & Reconstructed spatial positions\\
$\mathsf{S}^{(k)}$              & $\sR^{n^{(k)} \times m^{(k)} \times d_\mS}$           & Spatial positions random variable  \\
$\widetilde{\tS}^{(k)}$         & $\sR^{n^{(k)} \times m^{(k)} \times d}$               & Spatial positional embeddings  \\
$\mR^{(k)}$                     & $\sR^{n^{(k)} \times  d_\mR}$                         & Per-sensor calibration tokens \\
$\tR^{(k)}$                     & $\sR^{n^{(k)} \times m^{(k)} \times  d_\mR}$          & Per-sensor calibration tokens (broadcasted) \\
$\mathsf{R}^{(k)}$              & $\sR^{n^{(k)} \times m^{(k)} \times  d_\mR}$          &  Per-sensor calibration tokens, random variable \\
$\widetilde{\tR}^{(k)}$         & $\sR^{n^{(k)} \times m^{(k)} \times  d}$              &  Per-sensor calibration embeddings \\
$\mM^{(k)}$                     & $\{0, 1\}^{n^{(k)}\times m^{(k)}}$                    & Mask \\
$\overline{\mM}^{(k)}$          & $\{0, 1\}^{n^{(k)}\times m^{(k)}}$                    & Complement mask \\
$\mO^{(k)}$                & $\{0, 1\}^{n^{(k)}\times m^{(k)}}$             & Observation mask ($1$: observed) \\
$\widetilde{\mZ}^{(k)}$         & $\sR^{m_\mM^{(k)} \times d}$                          & Per-modality encoder output (pre-fusion) \\
$\mZ$                           & $\sR^{(\sum_k m_\mM^{(k)}) \times d}$                 & Scene representation (joint, post-fusion) \\
$\mathsf{Z}$                    & $\sR^{(\sum_k m_\mM^{(k)}) \times d}$                 & Scene representation, random variable \\
\bottomrule
\end{tabular}
\end{center}
\vspace{-0.2in}
\end{table}

\section*{Appendix 4: Confusion Matrices}
\begin{figure}[H]
    \centering
    \includegraphics[width=0.75\linewidth]{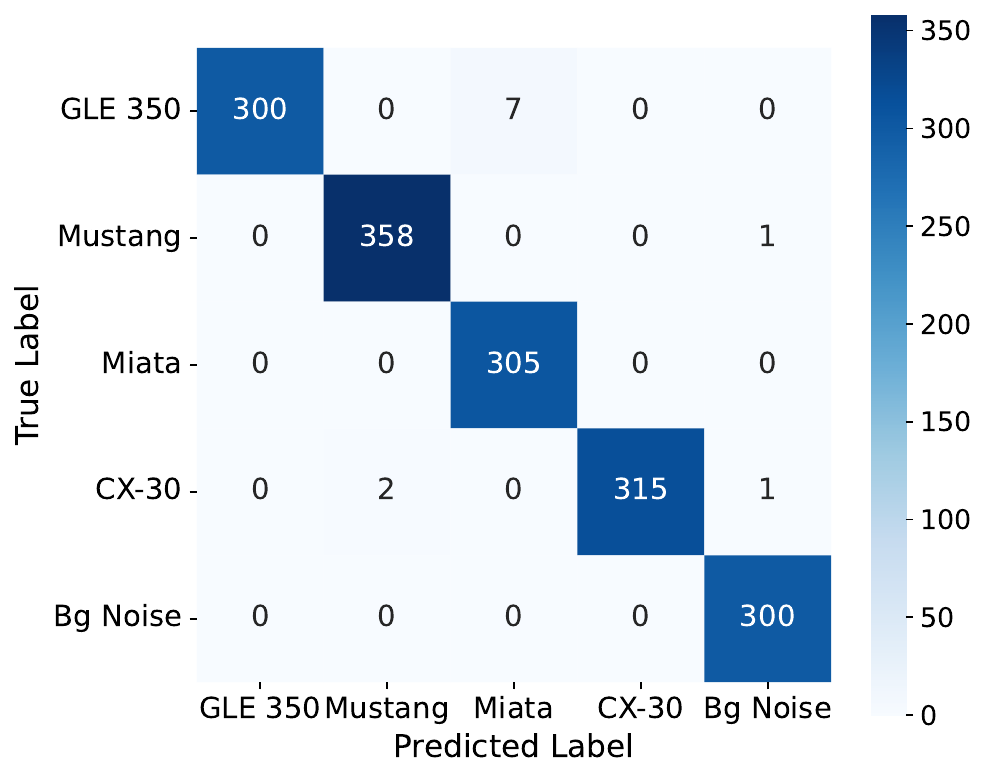}
    \vspace{-0.2in}
    \caption{\small Confusion matrix of \MODEL{} on the M3N-VC single-vehicle recognition task. Most classes are clearly separated, and the residual confusion aligns with the conceptual closeness of the vehicle classes.}
    \label{fig:cls_vehicle}
    \vspace{-0.15in}
\end{figure}

\begin{figure}[H]
    \centering
    \includegraphics[width=0.95\linewidth]{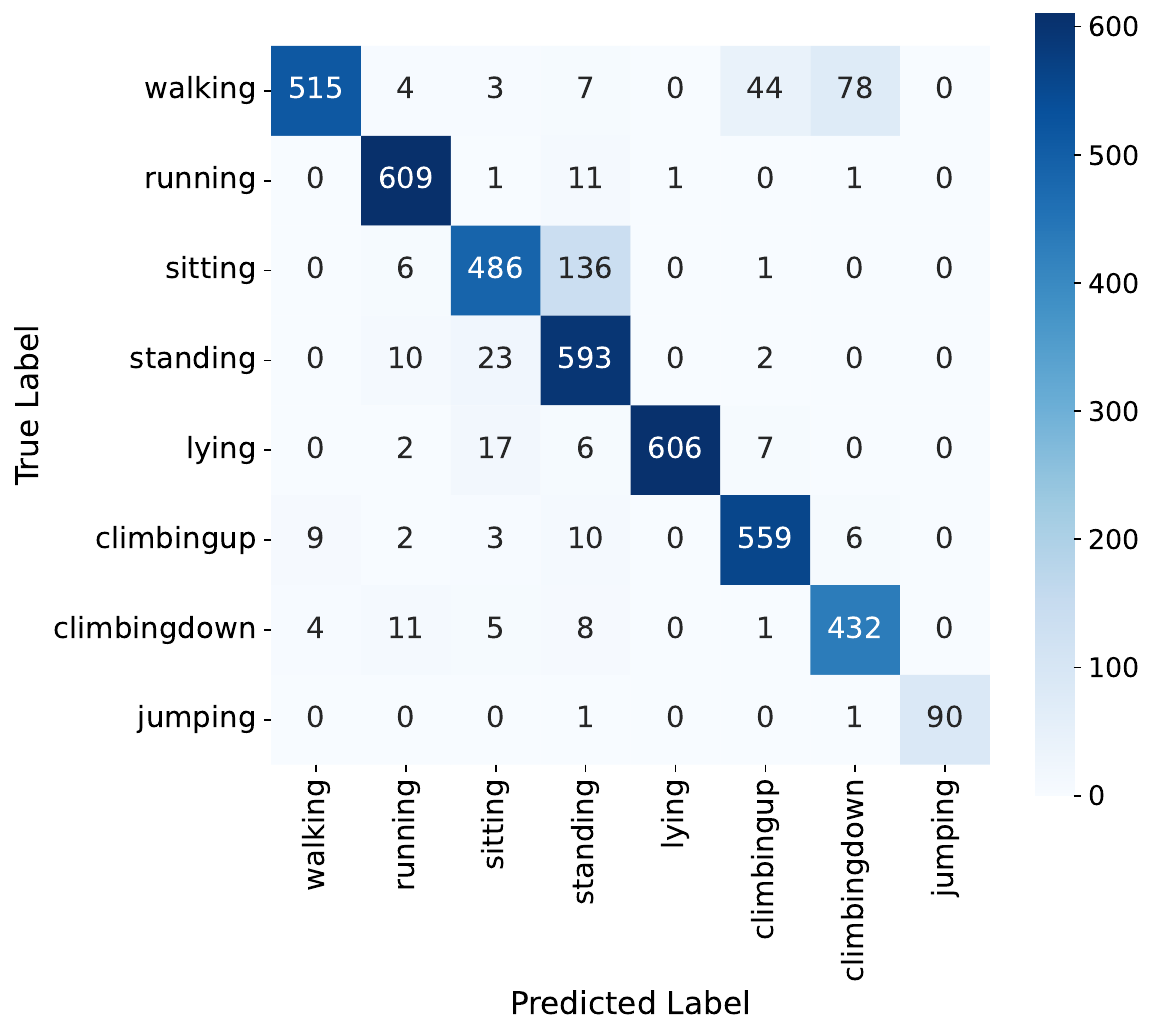}
    \vspace{-0.2in}
    \caption{\small Confusion matrix of \MODEL{} on the RealWorld-HAR activity recognition task. Most activities are clearly separated; the limited residual confusion is concentrated among activities with similar motion signatures.}
    \label{fig:cls_activity}
    \vspace{-0.2in}
\end{figure}

Figures~\ref{fig:cls_vehicle} and~\ref{fig:cls_activity} show the confusion matrices of \MODEL{} on the M3N-VC single-vehicle recognition task (Section~\ref{exp:m3n-vc}) and the RealWorld-HAR activity recognition task (Section~\ref{sec:other_apps}).

\end{document}